\documentclass[10pt,twocolumn,letterpaper]{article}

\usepackage[pagenumbers]{cvpr} % To force page numbers, e.g. for an arXiv version

\usepackage{graphicx}
\usepackage{amsmath}
\usepackage{amssymb}
\usepackage{booktabs}
\usepackage{booktabs}
\usepackage{multirow}
\usepackage{amsmath,amssymb}

\usepackage[pagebackref,breaklinks,colorlinks]{hyperref}

\usepackage[capitalize]{cleveref}
\crefname{section}{Sec.}{Secs.}
\Crefname{section}{Section}{Sections}
\Crefname{table}{Table}{Tables}
\crefname{table}{Tab.}{Tabs.}

\def\cvprPaperID{*****} % *** Enter the CVPR Paper ID here
\def\confName{CVPR}
\def\confYear{2022}

\begin{document}

\title{Next-ViT: Next Generation Vision Transformer for Efficient Deployment in Realistic Industrial Scenarios}

% \and

\author{%
    Jiashi Li\footnotemark[1],\qquad Xin Xia{\footnotemark[1]   \footnotemark[2]},\qquad Wei Li\footnotemark[1],\qquad Huixia Li, \qquad Xing Wang, \\
    {Xuefeng Xiao,\qquad Rui Wang, \qquad Min Zheng,\qquad Xin Pan}\\
 ByteDance Inc \\
 	{\tt\small \{lijiashi, xiaxin.97, liwei.97, lihuixia, wangxing.613\}@bytedance.com}\\
 	 	{\tt\small \{xiaoxuefeng.ailab, ruiwang.rw, zhengmin.666, panxin.321\}@bytedance.com}}

\maketitle

%%%%%%%%% ABSTRACT
\renewcommand{\thefootnote}{\fnsymbol{footnote}} %将脚注符号设置为fnsymbol类型，即特殊符号表示
\footnotetext[1]{Equal contribution.}
\footnotetext[2]{Corresponding author.}

\begin{abstract}
Due to the complex attention mechanisms and model design, most existing vision Transformers (ViTs) can not perform as efficiently as convolutional neural networks (CNNs) in realistic industrial deployment scenarios, e.g. TensorRT and CoreML. This poses a distinct challenge: 
\emph{Can a visual neural network be designed to infer as fast as CNNs and perform as powerful as ViTs?} In these work, we propose a next generation vision Transformer for efficient deployment in realistic industrial scenarios, namely Next-ViT, which dominates both CNNs and ViTs from the perspective of latency/accuracy trade-off. The Next Convolution Block (NCB) and Next Transformer Block (NTB) are respectively developed to capture local and global information with deployment-friendly mechanisms. Then, Next Hybrid Strategy (NHS) is designed to stack NCB and NTB in an efficient hybrid paradigm, which boosts performance in various downstream tasks. Extensive experiments show that Next-ViT significantly outperforms existing CNNs, ViTs and CNN-Transformer hybrid architectures with respect to the latency/accuracy trade-off across various vision tasks. On TensorRT, Next-ViT surpasses ResNet by \textbf{5.5} mAP (from 40.4 to 45.9) on COCO detection and \textbf{7.7\%} mIoU (from 38.8\% to 46.5\%) on ADE20K segmentation under similar latency. Meanwhile, it achieves comparable performance with CSWin, while the inference speed is accelerated by \textbf{3.6×}. On CoreML, Next-ViT surpasses EfficientFormer by \textbf{4.6} mAP (from 42.6 to 47.2) on COCO detection and \textbf{3.5\%} mIoU (from 45.1\% to 48.6\%) on ADE20K segmentation under similar latency. Our code and models are made public at:\url{https://github.com/bytedance/Next-ViT}
\end{abstract}

\section{Introduction}

\begin{figure*} [t]
\centering
\subfloat [ImageNet-1K classification on TensorRT] {\includegraphics[width=0.32\textwidth]{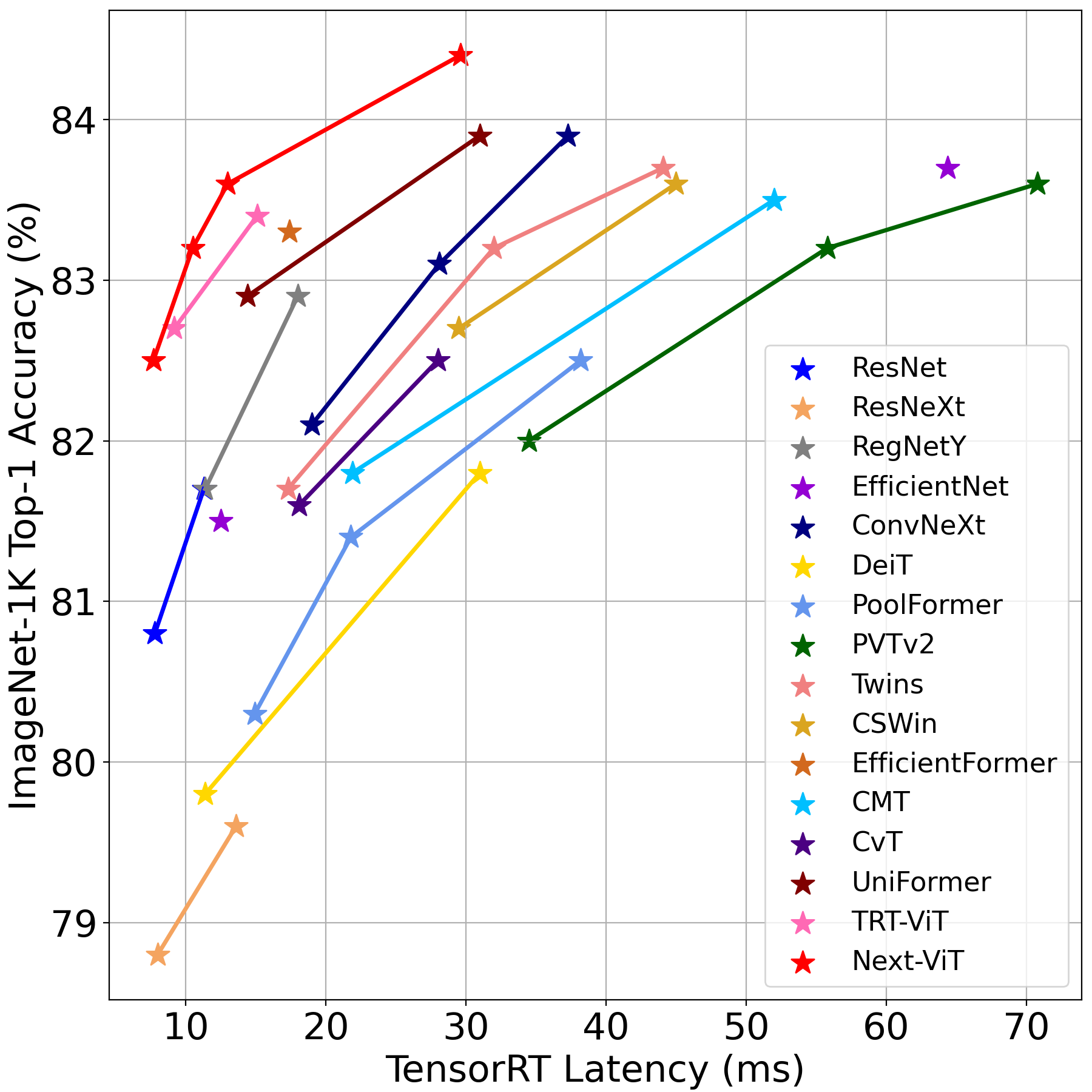}\label{cls_trt}}
\hspace{0.1mm}
\subfloat [COCO detection on TensorRT] {\includegraphics[width=0.32\textwidth]{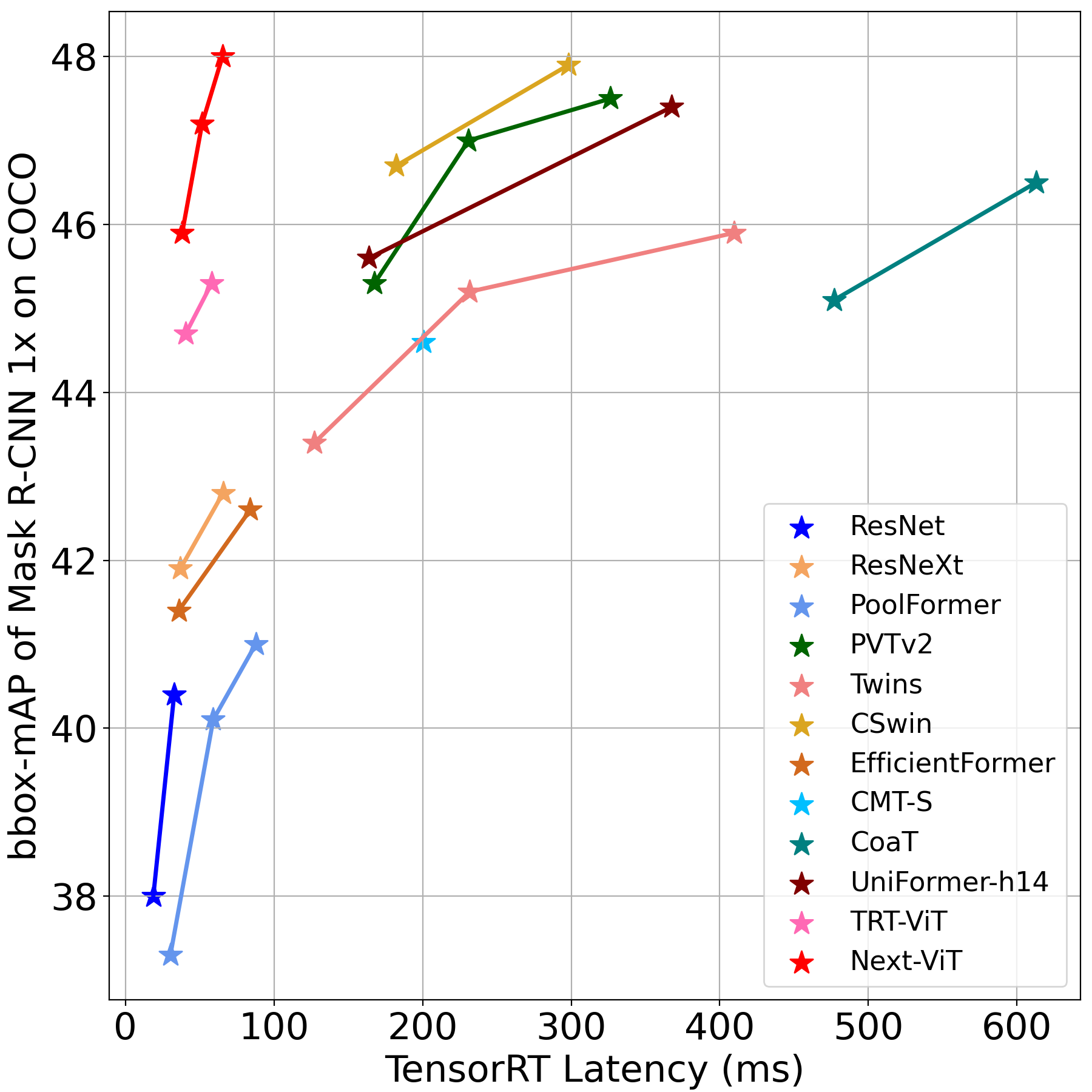}\label{det_trt}}
\hspace{0.1mm}
\subfloat [ADE20K segmentation on TensorRT] {\includegraphics[width=0.32\textwidth]{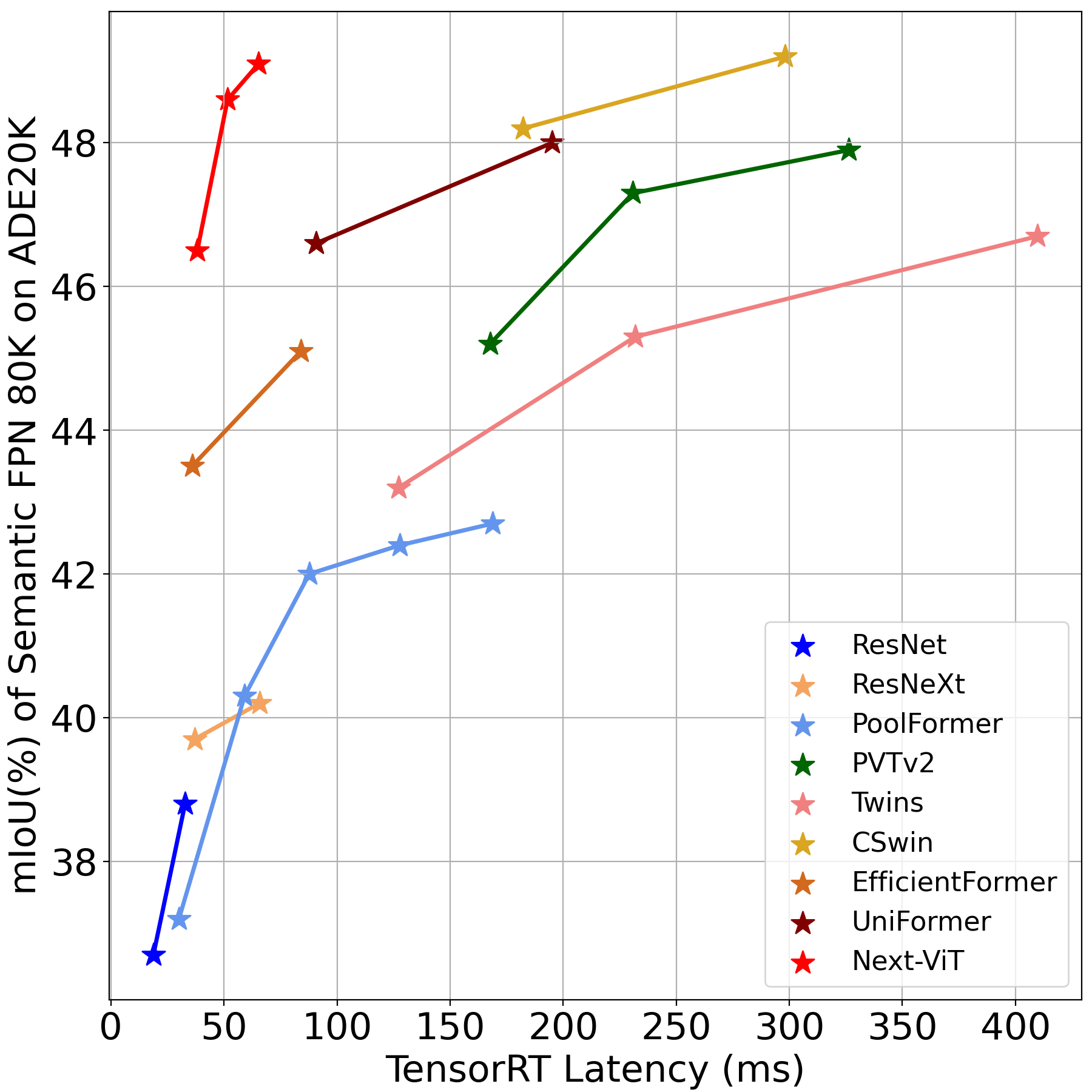}\label{seg_trt}}
\hfil
\subfloat [ImageNet-1K classification on CoreML] {\includegraphics[width=0.32\textwidth]{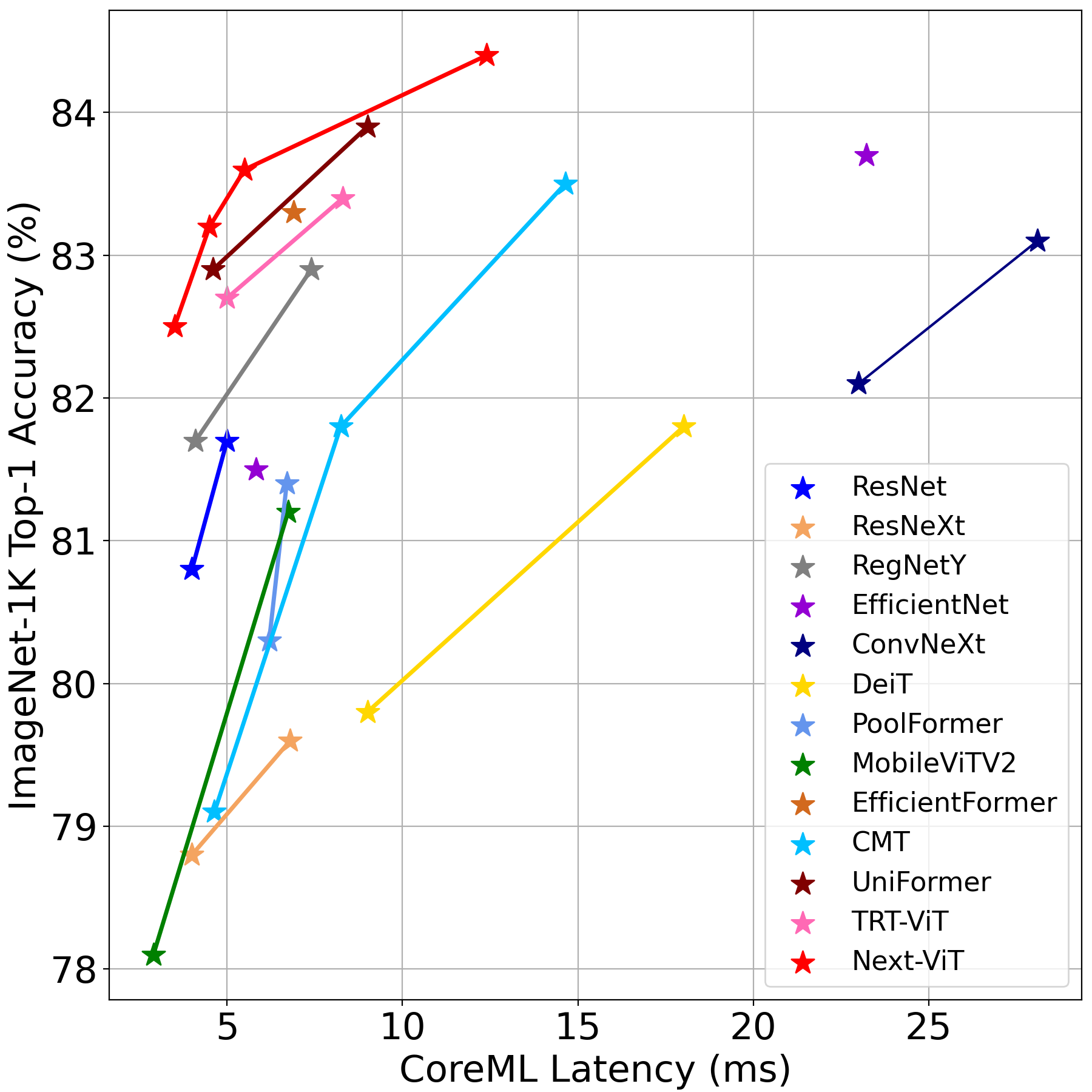}\label{cls_coreml}}
\hspace{0.1mm}
\subfloat [COCO detection on CoreML] {\includegraphics[width=0.32\textwidth]{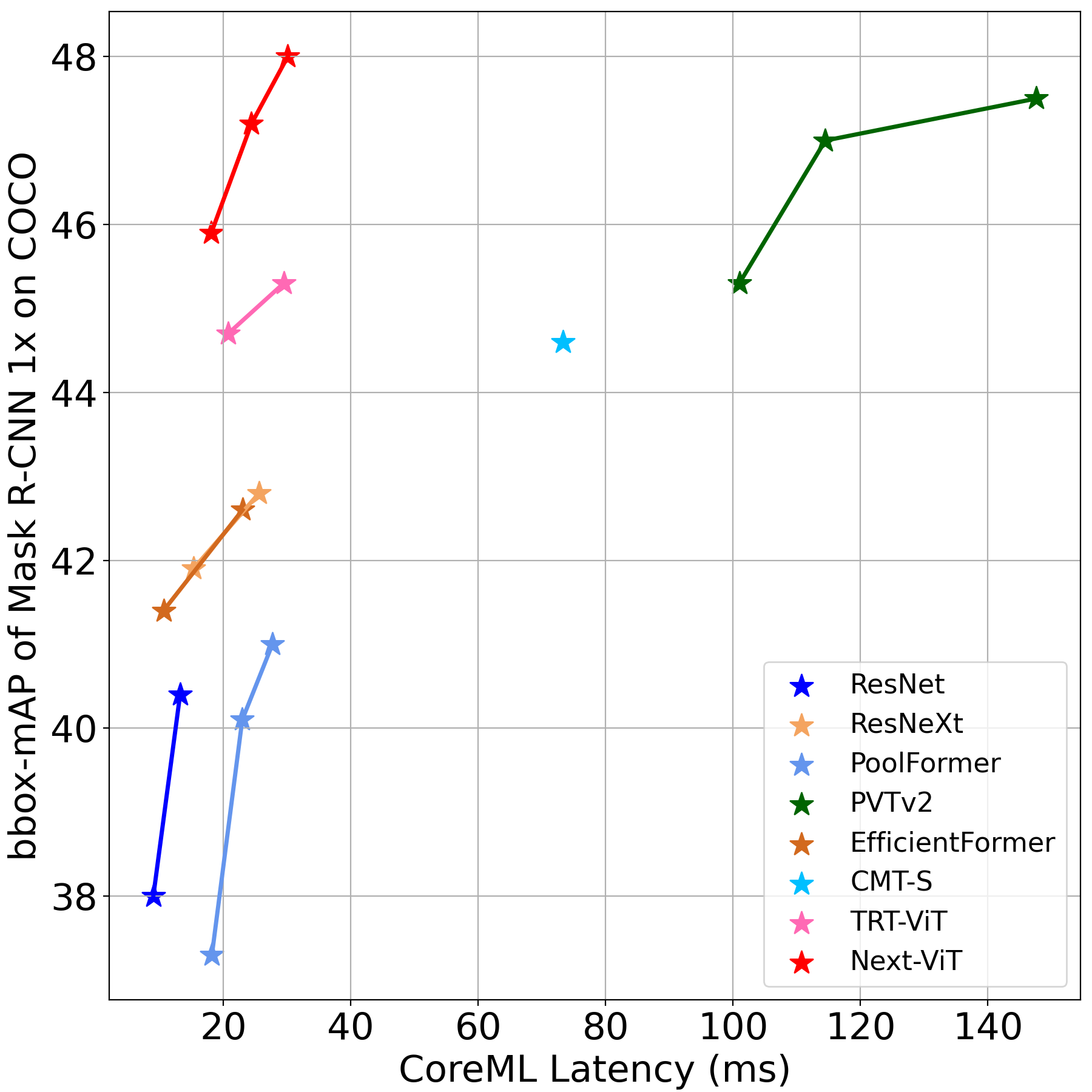}\label{det_coreml}}
\hspace{0.1mm}
\subfloat [ADE20K segmentation on CoreML] {\includegraphics[width=0.32\textwidth]{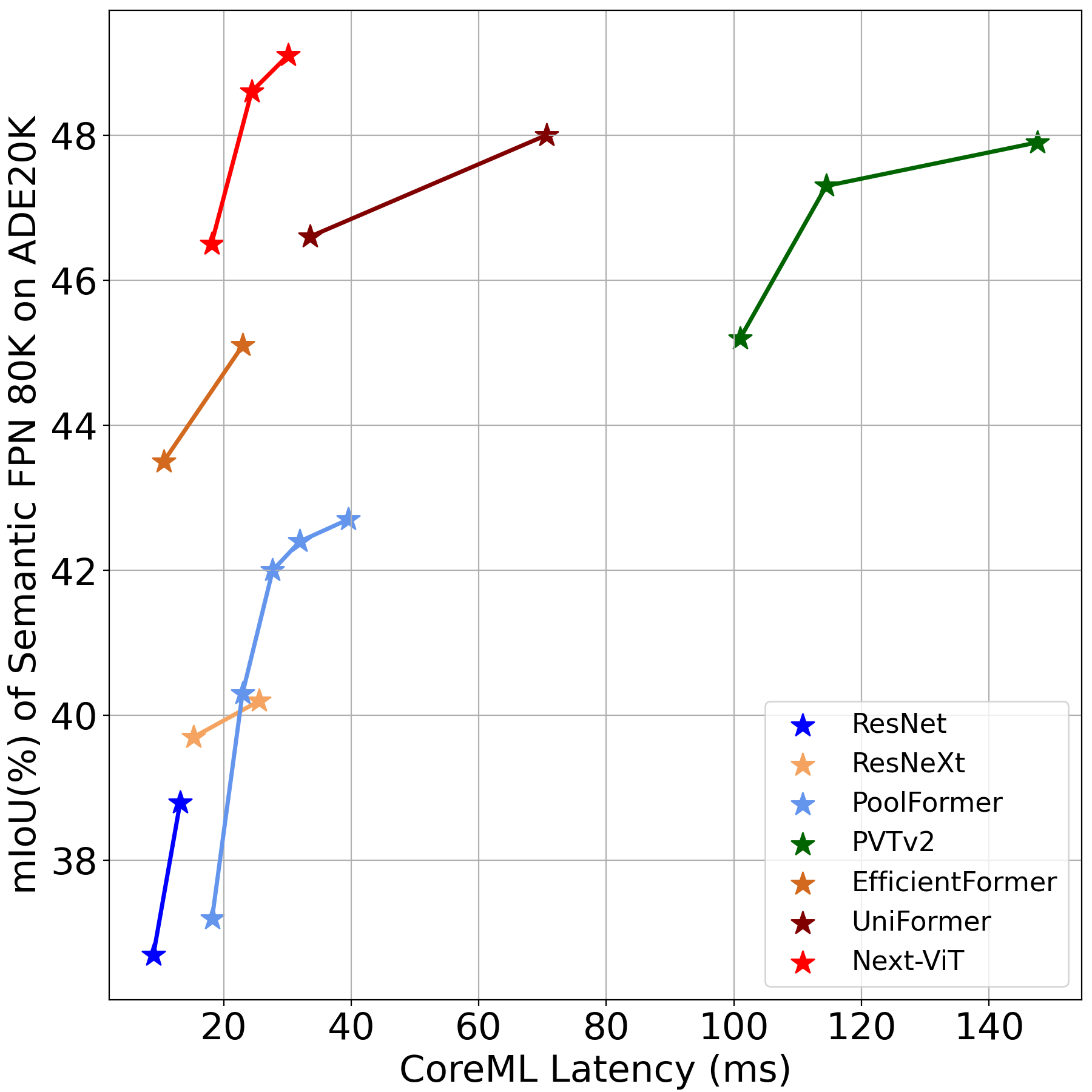}\label{seg_coreml}}

\caption{Comparison among Next-ViT and efficient Networks, in terms of accuracy-latency trade-off.}
\label{fig:accuracy}
\end{figure*}

Recently, vision Transformers (ViTs) have received increasing attention in industry and academia, and demonstrated much success in various computer vision tasks, such as image classification, object detection, semantic segmentation and etc. However, CNNs still dominate vision tasks from a real-world deployment perspective, because ViTs are usually much slower than classical CNNs, e.g. ResNets. There are some factors that limit the inference speed of the Transformer model, including quadratic complexity with respect to token length of the Multi-Head Self Attention (MHSA) mechanism, non-foldable LayerNorm and GELU layers, the complex model design causes frequent memory access and copying, etc.

Many works have struggled to free ViTs from high latency dilemma. For example, Swin Transformer \cite{Swin} and PVT \cite{PVT_v1} try to design more efficient spatial attention mechanisms to alleviate the quadratic-increasing computation complexity of MHSA. The others \cite{CoAtNet, li2022efficientformer, MobileViT} consider combining efficient convolution blocks and powerful Transformer blocks to design CNN-Transformer hybrid architecture to obtain a better trade-off between accuracy and latency. Coincidentally, almost all existing hybrid architectures \cite{CoAtNet, li2022efficientformer, MobileViT} adopt convolution blocks in the shallow stages and just stack Transformer block in the last few stages. However, we observe that such a hybrid strategy is effortless to lead to performance saturation on downstream tasks (e.g. segmentation and detection). Furthermore, we found that both convolution blocks and Transformer blocks in existing works can not possess characteristics of efficiency and performance at the same time. Although the accuracy-latency trade-off has been improved when compared with Vision Transformer, the overall performance of the existing hybrid architecture is still far away from satisfactory.
% For instance, some efforts consider designing new architectures or operations by changing the linear layers with convolutional layers (CONV) , combining self-attention with MobileNet blocks, or introducing sparse attention, to reduce the computational cost, while other efforts leverage network searching algorithm or pruning to improve efficiency. Although the computation-performance trade-off has been improved by existing works, the fundamental question that relates to the applicability of Transformer models remains unanswered: Can powerful vision Transformers run at MobileNet speed and become a default option for edge applications? 

%Furthermore, relevant studies have shown that the Transformer block has a strong ability to capture low-frequency signals, including the global structure information of feature map, but the ability to capture high-frequency signals is insufficient. Local information can be captured by convolution in the local receptive field, so that high-frequency features can be effectively extracted by convolution block. 
%Moreover, we observed that the latency of the Transformer block is greatly affected by the number of channels and the size of feature map. These observations motivate us to raise a question: \textit{how to design a Vision Transformer that can capture high and low frequency signals at the same time, and inference as fast as CNNs?}
\begin{figure*}[]
    \centering
    \includegraphics[width=0.65\textwidth]{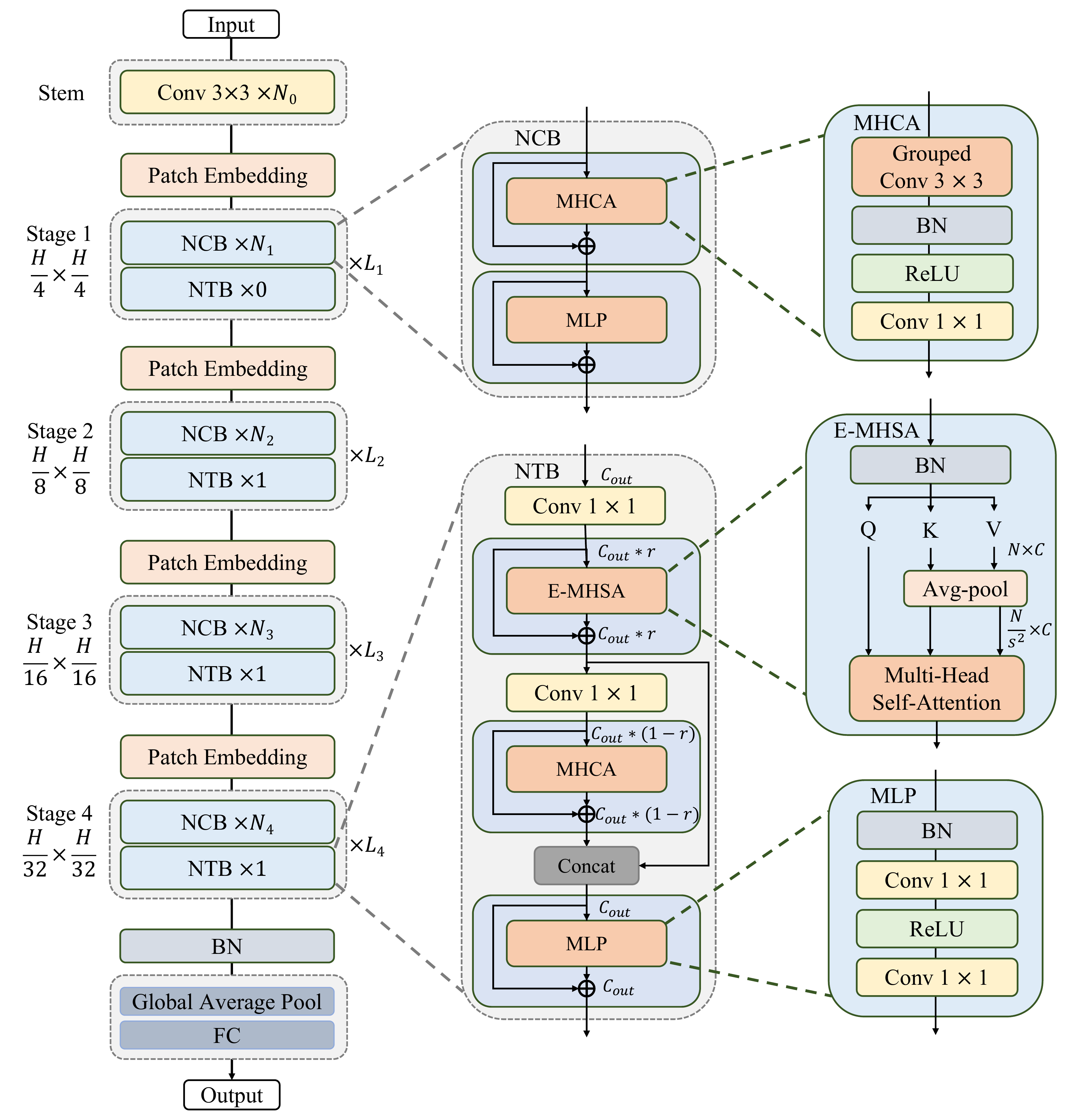}
    \caption{The left column is the overall hierarchical architecture of Next-ViT. 
    The medium column are the Next Convolution Block (NCB) and the Next Transformer Block (NTB). 
    The right column are the detailed visualization of multi-head convolutional attention (MHCA), efficient multi-head self-attention (E-MHSA) and the optimized MLP modules.}
    \label{fig:Next-ViT}
\end{figure*}

To address the above issues, this work develops three important components to design efficient vision Transformer networks. Firstly, we introduce the Next Convolution Block (NCB), which is skilled at capturing short-term dependency information in visual data with a novel deployment-friendly  Multi-Head Convolutional Attention (MHCA). Secondly, we build the Next Transformer Block (NTB), NTB is not only an expert in capturing long-term dependency information but also works as a lightweight and high-and-low-frequency signal mixer to enhance modeling capability. Finally, we design Next Hybrid Strategy (NHS) to stack NCB and NTB in a novel hybrid paradigm in each stage, which greatly reduce the proportion of the Transformer block and retaining the high precision of the vision Transformer network in various downstream tasks.
% Next Transformer Block first performs a channel dimension reduction on the input features. This is to reduce the computational and time-consuming overhead of global attention. The reduced dimensionality features go through a light-weight globol attention to capture low-frequency features and then extract high-frequency information through the previously designed local conv attention.  

% high-frequency information and further feature extraction through concat and MLP Layer operations. And like the next conv block, we use efficient bn and relu instead of layernorm and GELU to further reduce the inference speed. Through the above design, our Next Transformer Block can be both lightweight and capable of capturing high and low frequency signals at the same time.

Based on the above-proposed approaches, we propose next generation vision Transformer for realistic industrial deployment scenarios (abbreviated as Next-ViT). In this paper, to present a fair comparison, we provide a view that treats the latency on the specific hardware as direct efficiency feedback. TensorRT and CoreML represent generic and easy-to-deploy solutions for server-side and mobile-side devices, respectively, that help provide convincing hardware-oriented performance guidance. With this direct and accurate guidance, we redraw the accuracy and latency trade-off diagram of several existing competitive models in Figure \ref{fig:accuracy}. As depicted in Figure \ref{fig:accuracy}(a)(d), Next-ViT achieves best latency/accuracy trade-off on ImageNet-1K classification task. More importantly, Next-ViT shows a more significant latency/accuracy trade-off superiority on downstream tasks. As shown in Figure \ref{fig:accuracy}(b)(c), on TensorRT, Next-ViT outperforms ResNet by \textbf{5.5} mAP (from 40.4 to 45.9) on COCO detection and \textbf{7.7\%} mIoU (from 38.8\% to 46.5\%) on ADE20K segmentation under similar latency. Next-ViT achieves comparable performance with CSWin, while the inference speed is increased by \textbf{3.6×}. As depicted in Figure \ref{fig:accuracy}(e)(f), on CoreML, Next-ViT surpasses EfficientFormer by \textbf{4.6} mAP (from 42.6 to 47.2) on COCO detection and \textbf{3.5\%} mIoU (from 45.1\% to 48.6\%) on ADE20K segmentation under similar CoreML latency.

Our main contributions are summarized as follows:
\begin{itemize}
    \item We develop powerful convolution block and Transformer block, i.e. NCB and NTB, with deployment-friendly mechanisms. Next-ViT stacks NCB and NTB to build advanced CNN-Transformer hybrid architecture. 
    \item We design an innovative CNN-Transformer hybrid strategy from a new insight that boosts performance with high efficiency.
    \item We present Next-ViT, a family of powerful vision Transformer architecture. Extensive experiments demonstrate the advantage of Next-ViT. It achieves SOTA latency/accuracy trade-off on image classification, object detection and semantic segmentation on TensorRT and CoreML.
\end{itemize}

\begin{figure*} [t]
\centering
\subfloat [BottleNeck\cite{ResNet}] {\includegraphics[width=0.13\textwidth]{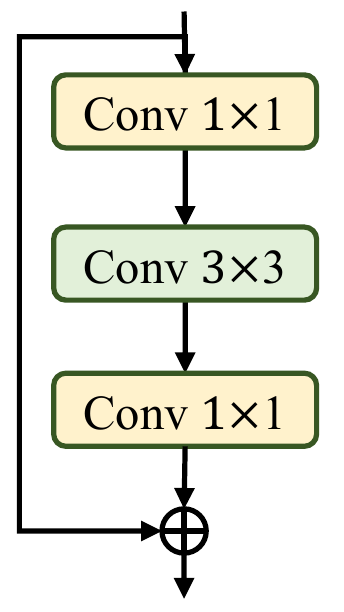}\label{block_resnet}}
\hspace{0.01mm}
\subfloat [ConvNeXt\cite{ConvNext}] {\includegraphics[width=0.13\textwidth]{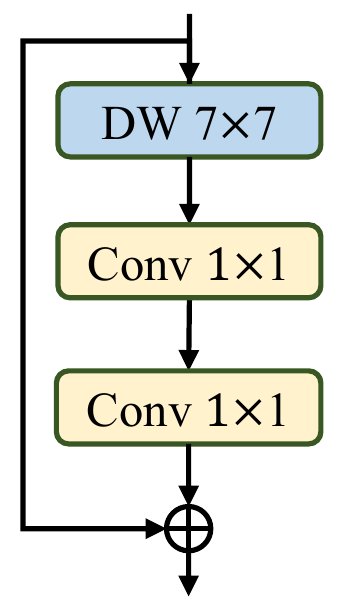}\label{block_convnext}}
\hspace{0.01mm}
\subfloat [Transformer\cite{ViT}] {\includegraphics[width=0.13\textwidth]{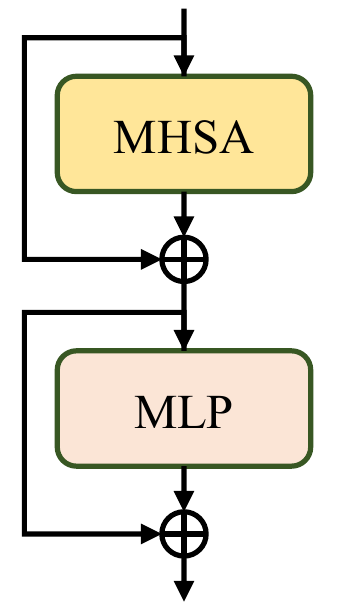}\label{block_mhsa}}
% \hfil
\hspace{0.01mm}
\subfloat [PoolFormer\cite{metaformer}] {\includegraphics[width=0.13\textwidth]{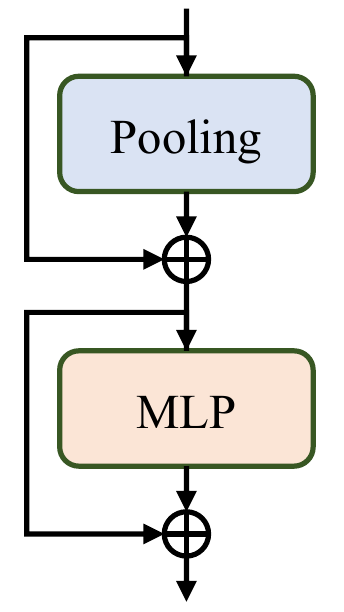}\label{block_poolformer}}
\hspace{0.01mm}
\subfloat [UniFormer\cite{uniformer}] {\includegraphics[width=0.13\textwidth]{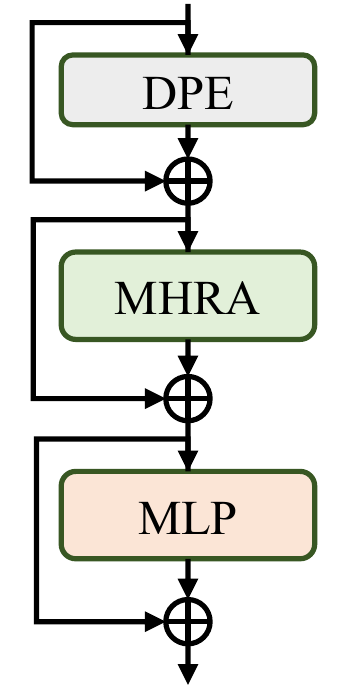}\label{block_uniformer}}
\hspace{0.01mm}
\subfloat [NCB (ours)] {\includegraphics[width=0.13\textwidth]{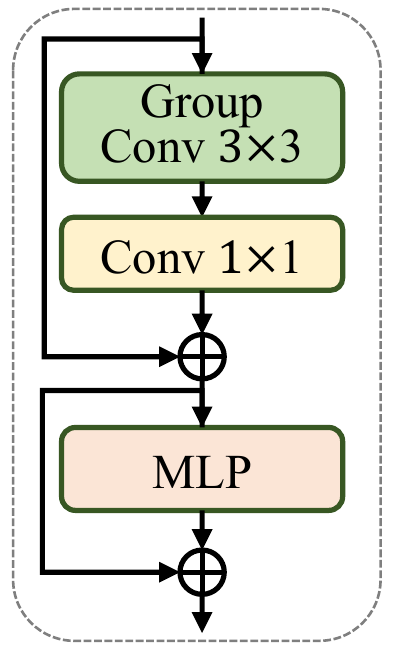}\label{fig:NCB}}
\hspace{0.01mm}
\subfloat [NTB (ours)] {\includegraphics[width=0.13\textwidth]{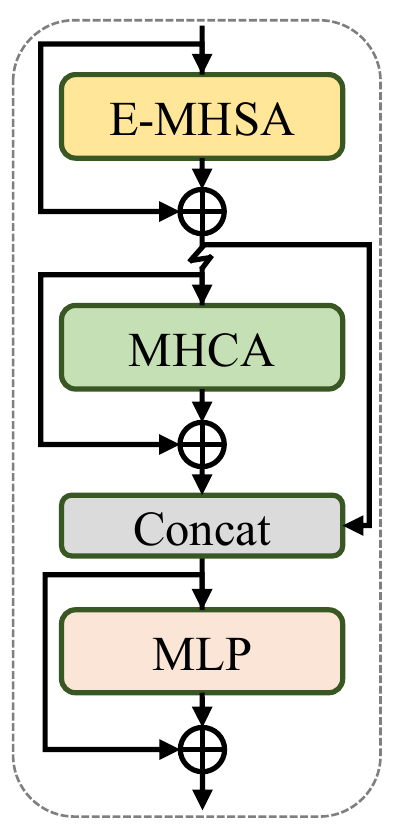}\label{fig:NTB}}
\caption{Comparison of different Transformer-based and convolution-based blocks.}
\label{fig:blocks}
\end{figure*}

\section{Related Work}
	\textbf{Convolutional Networks.}
    Over the past decade, Convolutional Neural Networks (CNNs) have dominated vision architectures in a variety of computer vision tasks, including image classification, object detection, and semantic segmentation.
    ResNet\cite{ResNet} uses residual connections to eliminate network degradation, ensuring that the network builds deeper and can capture high-level abstractions.
	DenseNet\cite{DenseNet} alternately enhances feature reuse and concatenates feature maps through dense connections.
	MobileNets\cite{Mobilenets, MobileNet_v2}  introduce depthwise convolution and point-wise convolution to build models with small memory and low latency.
	ShuffleNet\cite{Shufflenet} adopts group point-wise convolution and channel shuffle to reduce the computational cost further. ShuffleNetv2\cite{ShuffleNet_v2} propose that network architecture design should consider the direct metric such as speed, instead of the indirect metric like FLOPs. 
	ConvNeXt\cite{ConvNext} reviews the design of the vision Transformers and proposes a pure CNN model that can compete favorably with SOTA hierarchical vision Transformers across multiple computer vision benchmarks, while retaining the simplicity and efficiency of standard CNNs.
	
	\textbf{Vision Transformers.}
	Transformer is first proposed in the field of natural language processing (NLP). ViT\cite{ViT} splits the image into patches and treats these patches as words to perform self-attention, which shows that Transformer also achieves impressive performance on various vision tasks. DeiT\cite{Deit} introduces a teacher-student strategy specific to Transformers. T2T-ViT\cite{T2T} introduces a novel tokens-to-token (T2T) process to progressively tokenize images to tokens and structurally aggregate tokens. Swin Transformer\cite{Swin} proposes a general-purpose Transformer backbone, which constructs hierarchical feature maps and has linear computational complexity to image size. PiT\cite{PiT} incorporates a pooling layer into ViT, and shows that these advantages can be well harmonized to ViT through extensive experiments. Today, researchers pay more attention to efficiency, including efficient self-attention, training strategy, pyramid design, and etc.
	
	\textbf{Hybrid Models.}
	Recent works \cite{BoTNet,CvT,Cmt,MobileViT,li2022efficientformer, FAN} have shown that combining convolution and Transformer as a hybrid architecture helps absorb the strengths of both architectures.
	BoTNet\cite{BoTNet} replaces the spatial convolutions with global self-attention in the final three bottleneck blocks of ResNet. CvT\cite{CvT} introduces the depthwise and pointwise convolution in front of self-attention. CMT\cite{Cmt} proposes a new Transformer based hybrid network by taking advantage of Transformers to capture long-range dependencies and CNN to model local features.
	In MobileViT\cite{MobileViT}, introduces a light-weight and general- purpose vision Transformer for mobile devices. 
	Mobile-Former\cite{Mobile-Former} combines with the proposed lightweight cross attention to model the bridge, which is not only computationally efficient, but also has more representation power. EfficientFormer\cite{li2022efficientformer} complies with a dimension consistent design that smoothly leverages hardware-friendly 4D MetaBlocks and powerful 3D MHSA blocks. In this paper, we design a family of Next-ViT models that adapt more to the realistic industrial scenarios.

\section{Methods}
In this section, we first demonstrate the overview of the proposed Next-ViT.
Then, we discuss some core designs within Next-ViT, including the Next Convolution Block (NCB), Next Transformer Block (NTB) and the Next Hybrid Strategy (NHS). 
Moreover, we provide the architecture specifications with different model sizes.

\subsection{Overview}
% Building on the insight of ResNet \cite{ResNet} and recent ViTs \cite{Swin, Twins, CSWin}, 
We present the Next-ViT as illustrated in Figure \ref{fig:Next-ViT}.
By convention, Next-ViT follows the hierarchical pyramid architecture equipped with a patch embedding layer and a series of convolution or Transformer blocks in each stage.
The spatial resolution will be progressively reduced by 32$\times$ while the channel dimension will be expanded across different stages. 
In this chapter, we first dive deeper into designing the core blocks for information interaction and respectively develop powerful NCB and NTB to model short-term and long-term dependencies in visual data. The fusion of local and global information is also performed in NTB which further boosts modeling capability. Finally, we systematically study the manners of integrating convolution and Transformer blocks. To overcome the inherent defects of existing methods, we introduce Next Hybrid Strategy which stacks innovative NCB and NTB to build our advanced CNN-Transformer hybrid architecture.

\subsection{Next Convolution Block (NCB)}
To present the superiority of the proposed NCB, we first revisit some classical structural designs of convolution and Transformer blocks as shown in Figure \ref{fig:blocks}. BottleNeck block proposed by ResNet\cite{ResNet} has dominance in visual neural networks for a long time by its inherent inductive biases and deployment-friendly characteristics in most hardware platforms. Unfortunately, the effectiveness of the BottleNeck block is inadequate compared to the Transformer block. ConvNeXt block\cite{ConvNext} modernizes the BottleNeck block by imitating designs of Transformer block. While ConvNeXt block partly improves network performance, its inference speed on TensorRT/CoreML is severely limited by inefficient components, such as $7\times7$ depthwise convolution, LayerNorm, and GELU. Transformer block has achieved excellent results in the various visual task and its intrinsic superiority is jointly endowed by the paradigm of MetaFormer\cite{metaformer} and the attention-based token mixer module~\cite{Swin}~\cite{CSWin}. However, the inference speed of Transformer block is much slower than BottleNeck block due to its complex attention mechanisms, which is unbearable in most realistic industrial scenarios.

To overcome the defeats of the above blocks, we introduce a Next Convolution Block (NCB),  which maintains the deployment advantage of BottleNeck block while obtaining prominent performance as Transformer block. As shown in Figure \ref{fig:blocks}(f), NCB follows the general architecture of MetaFormer\cite{metaformer}, which is verified to be essential to the Transformer block. In the meantime, an efficient attention-based token mixer is equally important. We design a novel Multi-Head Convolutional Attention (MHCA) as an efficient token mixer with deployment-friendly convolution operation. Finally, we build NCB with MHCA and MLP layer in the paradigm of MetaFormer\cite{metaformer}. Our proposed NCB can be formulated as follows:
\begin{equation}
    \begin{aligned}
        \label{equ:ncb}
        & \tilde{z}^l = \text{MHCA}(z^{l-1}) + z^{l-1} \; \\
        & z^l = \text{MLP}(\tilde{z}^l) + \tilde{z}^l \;
    \end{aligned}
\end{equation}
where $z^{l-1}$ denotes the input from the $l-1$ block, $\tilde{z}^l$ and $z^l$ are the outputs of MHCA and the $l$ NCB. We will introduce MHCA in detail in the next section.

\subsubsection{Multi-Head Convolutional Attention (MHCA)}
To free the existing attention-based token mixer from the high latency dilemma, we design a novel attention mechanism with efficient convolution operation, i.e. Convolutional Attention (CA), for fast inference speed. In the meantime, inspired by the effective multi-head design in MHSA\cite{vaswani2017attention}, we build our convolutional attention with multi-head paradigm which jointly attend to information from different representation subspaces at a different position for effective local representation learning. The definition of proposed Multi-Head Convolutional Attention (MHCA) can be summarized as follows:
\begin{align}
    \label{equ:mhca}
    & \text{MHCA}(z) = \text{Concat}(\text{CA}_1(z_1),\text{CA}_2(z_2), ..., \text{CA}_h(z_h))W^P \;
    % & \text{Proj}(z) = \text{C}_1(\text{Act}(\text{Norm}(z)))  \; 
\end{align}
Here, MHCA captures information from $h$  parallel representation subspaces. $z = [z_1, z_2, ... , z_h]$ indicates to divide the input feature $z$ into multi-head form in channel dimension. To promote the information interaction across the multiple heads, we also equip MHCA with a projection layer ($W^P$).  CA is single-head convolutional attention which can be defined as:
\begin{equation}
    \begin{aligned}
        \label{equ:ca}
        % \text{CA}_i(z_i) = \text{Conv}(z_i \cdot W_i + bias) \;
        & \text{CA}(z) = \text{O}(W, (T_{m}, T_{n})) \ \ where \ \  T_{\{m,n\}} \ \ \in \ \ z \;
    \end{aligned}
\end{equation}
where $T_m$ and $T_n$ are adjacent tokens in input feature $z$. $O$ is an inner product operation with trainable parameter $W$ and input tokens $T_{\{m,n\}}$. CA is capable of learning affinity between different tokens in the local receptive field through iteratively optimizing trainable parameter $W$. Concretely, the implementation of MHCA is carried out with a group convolution (multi-head convolution) and a point-wise convolution, as shown in Figure \ref{fig:blocks}(f). We uniformly set head dim to 32 in all MHCA for fast inference speed with various date-type on TensorRT. Besides, we adopt efficient BatchNorm (BN) and ReLU activation function in NCB rather than LayerNorm (LN) and GELU in traditional Transformer blocks, which further accelerates inference speed. Experimental results in the ablation study show the superiority of NCB compared with existing blocks, e.g BottleNeck block, ConvNext block, LSA block and etc.

\begin{figure}[]
    \centering
    \includegraphics[width=0.47\textwidth]{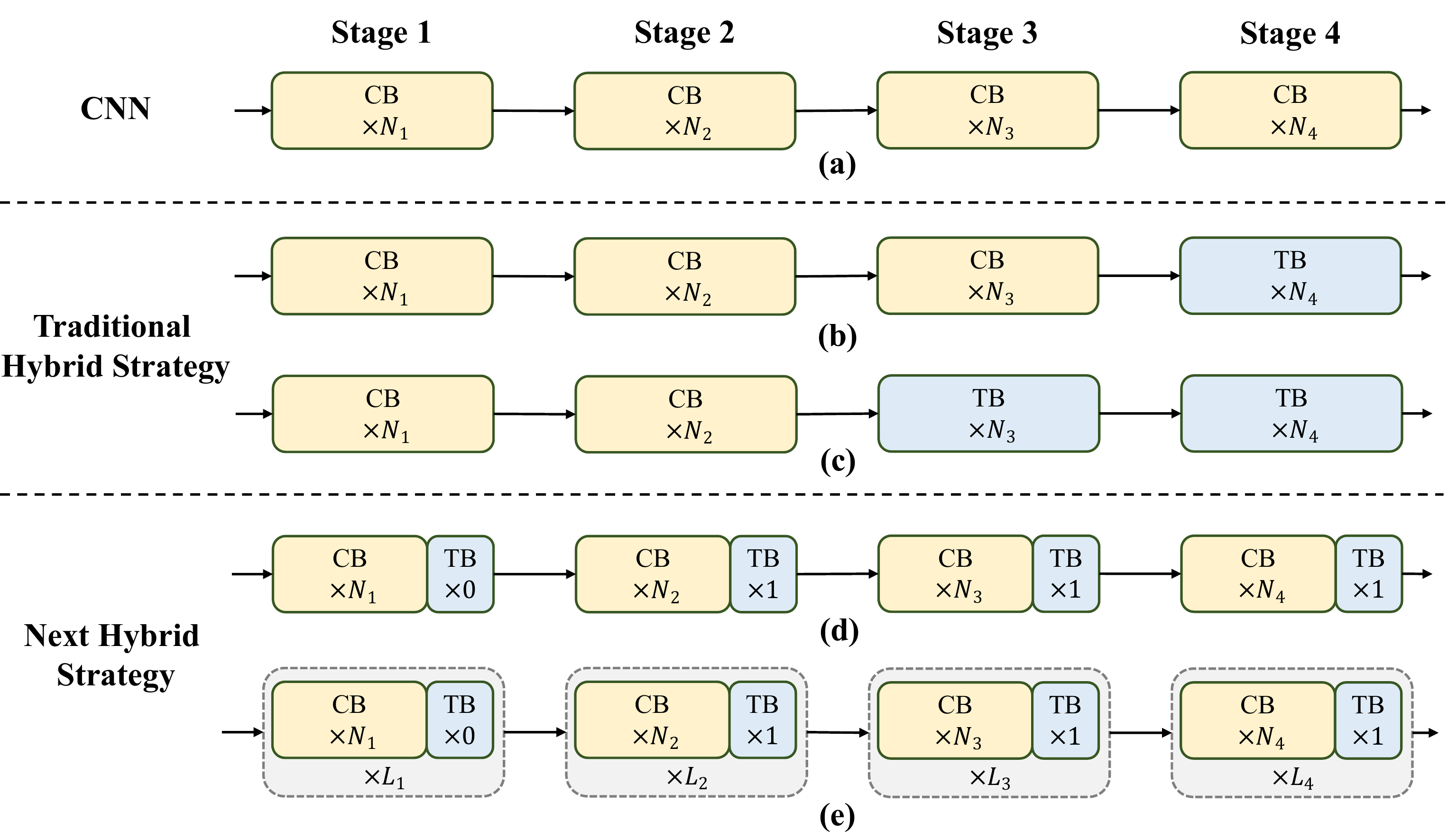}
    \caption{Comparison of traditional hybrid strategies and NHS.}
    \label{fig:NHS}
\end{figure}

\subsection{Next Transformer Block (NTB)}
Although the local representation has been effectively learned via NCB, the capture of global information is urgent to be addressed. Transformer block has a strong ability to capture low-frequency signals which provide global information (e.g global shapes and structures). Nevertheless, relevant studies \cite{park2022vision} have observed that Transformer blocks may deteriorate high-frequency information, such as local textures information, to a certain extent. Signals in different frequency segments are indispensable in the human visual system  ~\cite{bullier2001integrated}~\cite{kauffmann2014neural} and will be fused in some specific way to extract more essential and distinct features. 

Motivated by these observations, we develop the Next Transformer Block (NTB) to capture multi-frequency signals in the lightweight mechanism. Furthermore, NTB works as an effective multi-frequency signals mixer to further enhance overall modeling capability. As shown in Figure \ref{fig:Next-ViT}, NTB firstly captures low-frequency signals with an Efficient Multi-Head Self Attention(E-MHSA) which can be depicted as:
\begin{align}
    \label{equ:mhsa}
    \text{E-MHSA}(z) = \text{Concat}(\text{SA}_1(z_1), \text{SA}_2(z_2), ..., \text{SA}_h(z_h))W^P \;
\end{align}
where $z=[z_1, z_2, ..., z_h]$ denotes to divide the input feature $z$ into multi-head form in channel dimension. SA is a spatial reduction self-attention operator which is inspired by Linear SRA~\cite{PVT_v2} and performing as:
\begin{align}
    \label{equ:attn}
    \text{SA}(X) = \text{Attention}(X \cdot W^Q, \text{P}_s(X \cdot W^K), \text{P}_s(X \cdot W^V)) \;
\end{align}
where Attention represents a standard attention calculating as $\text{Attention}(Q, K, V) = \text{softmax}(\frac{QK^T}{d_k})V$, in which $d_k$ denotes the scaling factor. $W^Q, W^K, W^V$ are linear layers for context encoding. P$_s$ is an avg-pool operation with stride $s$ for downsampling the spatial dimension before the attention operation to reduce computational cost. Specifically, We observe the time consumption of the E-MHSA module is also greatly affected by its number of channels. NTB thus performs a channel dimension reduction before the E-MHSA module with point-wise convolutions to further accelerate inference. A shrinking ratio $r$ is introduced for channel reduction. We also utilize Batch Normalization in the E-MHSA module for extremely efficient deployment. 

Furthermore, NTB is equipped with an MHCA module that cooperates with the E-MHSA module to capture multi-frequency signals. After that, output features from E-MHSA and MHCA are concatenated to mix high-low-frequency information. Finally, an MLP layer is borrowed at the end to extract more essential and distinct features. Briefly, the implementation of NTB can be formulated as follows:
\begin{equation}
    \begin{aligned}
        \label{equ:mix}
        & \dot{z}^l = \text{Proj}(z^{l-1}) \; \\
        & \tilde{z}^l = \text{E-MHSA}(\dot{z}^l) + \dot{z}^l \; \\
        & \ddot{z}^l = \text{Proj}(\tilde{z}^l) \; \\
        & \hat{z}^l = \text{MHCA}(\ddot{z}^l) + \ddot{z}^l \; \\
        & \overline{z}^l = \text{Concat}(\tilde{z}^l , \hat{z}^l) \; \\
        & z^l = \text{MLP}(\overline{z}^l) + \overline{z}^l \;
    \end{aligned}
\end{equation}
where $\tilde{z}^l$, $\hat{z}^l$ and $z^l$ denote the output of E-MHSA, MHCA and NTB, respectively. Proj denotes the point-wise convolution layer for channel projection. Also, NTB uniformly adopts BN and ReLU as the efficient norm and activation layers instead of LN and GELU. Compared with traditional Transformer block, NTB is capable of capturing and mixing multi-frequency information in the lightweight mechanism which greatly boosts model performance. 

\subsection{Next Hybrid Strategy (NHS)}
Some recent works~\cite{CoAtNet, li2022efficientformer, MobileViT, xia2022trt} have paid great efforts to combine CNN and Transformer for efficient deployment. 
As shown in Figure \ref{fig:NHS}(b)(c), almost all of them monotonously adopt convolution blocks in the shallow stages and just stack Transformer blocks in the last one or two stages which presents effective results in the classification task. Unfortunately, we observe that these traditional hybrid strategies are effortless to reach performance saturation on downstream tasks (e.g. segmentation and detection). 
The reason is, the classification task just uses outputs from the last stage for prediction while downstream tasks (e.g. segmentation and detection) usually rely on features from each stage to gain better results. The traditional hybrid strategies, however, just stack Transformer blocks in the last few stages. The shallow stages thus fail to capture global information, e.g. global shapes and structures of object,  which is vital to segmentation and detection tasks.

\begin{table}[]
    \centering\caption{Comparison of different hybrid strategies. Cls denotes ImageNet-1K classification task. Det denotes detection task on COCO dataset with Mask-RCNN 1$\times$. Seg denotes segmentation task with Semantic FPN 80k on ADE20K dataset. TensorRT latency is uniformly measured with the input size of $8\times 3 \times 224 \times 224$.}
    \resizebox{0.35\textwidth}{!}{
        \begin{tabular}{c|c|c|c|c}
            \toprule
            \multirow{2.5}{*}{Model}      & Latency(ms)       & Cls   & Det      & Seg      \\ \cmidrule(l){2-5}
                                                    & TensorRT              & Acc(\%)       & AP$^b$                        & mIoU(\%)    \\ \midrule
            ResNet101                               & 7.8               & 80.8          & 40.4                          & 38.8          \\ \midrule
            C C C C                                 & 7.8               & 82.1          & 44.1                          & 43.2          \\ 
            C C C T                                 & 7.8               & 82.4          & 44.7                          & 45.2          \\ 
            C C T T                                 & 7.8               & 81.5          & 44.0                          & 44.8          \\ 
            C T T T                                 & 8.0               & 80.6          & 42.6                          & 44.1          \\ \midrule
            C C C H$_\text{N}$                                     & 7.7               & 82.1          & 44.6           & 44.8          \\ 
            C C H$_\text{N}$ H$_\text{N}$                         & 7.7               & 82.3          & 45.2              & 45.8          \\ 
            \textbf{C H$_\text{N}$ H$_\text{N}$ H$_\text{N}$ }      & \textbf{7.7}  & \textbf{82.3}  & \textbf{45.5}       & \textbf{46.0}          \\ 
            H$_\text{N}$ H$_\text{N}$ H$_\text{N}$ H$_\text{N}$       & 7.8               & 81.7          & 44.5                          & 45.3          \\
            \bottomrule                          
    \end{tabular}}
    \label{tab:NHS1}
\end{table}
\begin{table}[]
    \centering\caption{Comparison of different patterns in NHS and exploration of different hyper-parameters configurations. S1,S2,S3 and S4 denote Stage1, Stage2, Stage3 and Stage4 respectively.}
    \resizebox{0.48\textwidth}{!}{
        \begin{tabular}{c|c|c|c|c|c|c|c|c|c}
            \toprule
            \multirow{2.5}{*}{Hybrid Strategy}       &\multirow{2.5}{*}{Model}            & \multicolumn{4}{c|}{Configs}              & Latency(ms)      & Cls   & Det     & Seg       \\ \cmidrule(l){3-10}
                                                     &                                    & S1 & S2 & S3 & S4 & TensorRT  & Acc(\%)     & AP$^b$  & mIoU(\%)    \\ \midrule
            \multirow{3}{*}{$\begin{bmatrix} \text{NCB} \times N \\ \text{NTB} \times 1 \end{bmatrix}$}   
            & Small   & 3 & 4 & (11+1) & 3             & 7.7           & 82.3         & 45.5                          & 46.0          \\ 
            & Base    & 3 & 4 & (22+1) & 3             & 9.7           & 83.0          & 45.8                          & 46.7          \\ 
            & Large   & 3 & 4 & (33+1) & 3             & 11.7           & 83.2          & 46.2                          & 47.0          \\ \midrule
            \multirow{7}{*}{$\begin{bmatrix} \text{NCB} \times N \\ \text{NTB} \times 1 \end{bmatrix} \times L$}    & \multirow{5}{*}{Small}   & 3 & 4 & (1+1)$\times$3 & 3            & 7.4           & 82.1          & 44.9                          & 45.7          \\ 
                        &                           & 3 & 4 & (2+1)$\times$3 & 3       & 8.0           & 82.5          & 45.6                          & 45.8          \\ 
                        &                           & 3 & 4 & (3+1)$\times$3 & 3       & 8.5           & 82.6          & 46.1                          & 46.7          \\ 
                        &                           & \textbf{3} & \textbf{4} & \textbf{(4+1)$\times$2} & \textbf{3}       &\textbf{7.7}   &\textbf{82.5}  & \textbf{45.9}        & \textbf{46.5}          \\ 
                        &                           & 3 & 4 & (5+1)$\times$2 & 3       & 8.1           & 82.5          & 45.9                          & 46.6          \\ \cmidrule(l){2-10} 
            & \textbf{Base}   & \textbf{3} & \textbf{4} & \textbf{(4+1)$\times$4} & \textbf{3}       & \textbf{10.5}           & \textbf{83.2}          & \textbf{47.2}                          & \textbf{48.6}          \\ 
            & \textbf{Large}  & \textbf{3} & \textbf{4} & \textbf{(4+1)$\times$6} & \textbf{3}       & \textbf{13.0}           & \textbf{83.6}          & \textbf{48.0}                           & \textbf{49.1}          \\ 
            \bottomrule                          
    \end{tabular}}
    \label{tab:NHS2}
\vspace{-0.25cm}
\end{table}
To overcome the defeats of existing hybrid strategies, we propose a Next Hybrid Strategy (NHS) from a new insight, which creatively stacks convolution block (NCB) and Transformer block (NTB) with $(N+1)*L$ hybrid paradigm. NHS significantly promotes model performance in downstream tasks under controlling the proportion of Transformer block for efficient deployment.
Firstly, in order to endow the shallow stages with the capability of capturing global information, we present a novel hybrid strategy in $(\text{NCB} \times N + \text{NTB} \times 1)$ pattern, which sequentially stack $N$ NCB and one NTB in each stage as shown in Figure \ref{fig:NHS}(d). Specifically, the Transformer block(NTB) is placed at the end of each stage, which enables the model to learn global representation in the shallow layers. 
We conduct a series of experiments to verify the superiority of the proposed hybrid strategy. The performances of difference hybrid strategies are shown in Table \ref{tab:NHS1}. $\text{C}$ denotes uniformly stacking convolution block(NCB) in one stage and $\text{T}$ denotes consistently building one stage with Transformer block(NTB). Specially, $\text{H}_\text{N}$ indicates stacking NCB and NTB with $(\text{NCB} \times N + \text{NTB} \times 1)$ pattern in the corresponding stage. All models in Table~\ref{tab:NHS1} are equipped with four stages. For example, $\text{C}\,\text{C}\,\text{C}\,\text{C}$ represents consistently using convolution block in all of the four stages. For fair comparison, we build all the model under similar TensorRT latency. More implementation details are presented in Section~\ref{Experimental_section}.
As shown in Table~\ref{tab:NHS1}, the proposed hybrid strategy significantly promotes model performance compared with existing methods in the downstream tasks. $\text{C}\,\text{H}_\text{N}\,\text{H}_\text{N}\,\text{H}_\text{N}$ achieve the best overall performance. For example, $\text{C}\,\text{H}_\text{N}\,\text{H}_\text{N}\,\text{H}_\text{N}$ surpasses $\text{C}\,\text{C}\,\text{C}\,\text{T}$ 0.8 mAP in detection and 0.8\% mIoU in segmentation. Besides, The results of $\text{H}_\text{N}\,\text{H}_\text{N}\,\text{H}_\text{N}\,\text{H}_\text{N}$ shows that placing Transformer block in the first stage will deteriorate the latency-accuracy trade-off of model.

We further verify the general effectiveness of $\text{C}\,\text{H}_\text{N}\,\text{H}_\text{N}\,\text{H}_\text{N}$ on large model by increasing the number of blocks in the third stage as ResNet~\cite{ResNet}. Experimental results of the first three rows in Table~\ref{tab:NHS2} show that the performance of large model is hard to promote and gradually reaches saturation.
Such a phenomenon indicates that expanding model size by enlarging the $N$ of $(\text{NCB} \times N + \text{NTB} \times 1)$ pattern, i.e. simply adding more convolution block is not the best choice. It also implies that the value of $N$ in $(\text{NCB} \times N + \text{NTB} \times 1)$ pattern may seriously affect the model performance.
% Through analysis, we discover the low-frequency information will be gradually erased with the depth of NCB.
We thus begin to explore the impact of the value of $N$ on the model performance through extensive experiments. As shown in Table~\ref{tab:NHS2} (middle), we build models with different configurations of $N$ on the third stage. To build model with similar latency for fair comparison, we stack $L$ groups of $(\text{NCB} \times N + \text{NTB} \times 1)$ pattern when the value of $N$ is small. 
Surprisingly, we found that stack NCB and NTB in $(\text{NCB} \times N + \text{NTB} \times 1)\times L$ pattern achieve better model performance compared to $(\text{NCB} \times N + \text{NTB} \times 1)$ pattern.
It denotes that repeatedly combining low-frequency signal extractors and high-frequency signal extractors in a proper manner($(\text{NCB} \times N + \text{NTB} \times 1)$) leads to higher quality representation learning.  As shown in Table~\ref{tab:NHS2}, model with $N=4$ in the third stage achieving the best trade-off between performance and latency.
We further build the larger model by enlarging $L$ of  $(\text{NCB} \times 4 + \text{NTB} \times 1)\times L$ pattern in the third stage. 
As shown in Table~\ref{tab:NHS2} (bottom), the performance of Base ($L=4$) and Large ($L=6$) model are significantly promote compared to small model, which verifies the general effectiveness of proposed $(\text{NCB} \times N + \text{NTB} \times 1)\times L$ pattern. We use $N=4$ as the basic configurations in the rest of the paper.
% To verify the hypothesis, we conduct a series of experiments and the results are reported in Table~\ref{tab:NHS2}. As we can see from Table~\ref{tab:NHS2}, $(\text{NCB} \times N + \text{NTB} \times 1)\times L$ hybrid strategy breaks the previous performance bottleneck. Specifically, the large model achieves 50.4\% mIoU on Segmentation task, which is 2.9\% higher than previous large model, which reveals that $(\text{NCB} \times N + \text{NTB} \times 1)\times L$ pattern achieves excellent potential
% on downstream tasks.

We stack NCB and NTB with the above Next Hybrid Strategy to build Next-ViT, which can be formally defined as:
\begin{align}
    \label{equ:nhs}
    \text{Next-ViT}(X) = \oint_i \{[\varGamma(\varPsi (X) \times N_i)] \times L_i \}\;
\end{align}
where $i \in{(1,2,3,4)}$ denotes the stage index. $\varPsi$ means NCB. $\varGamma$ means identity layer when $i=1$, otherwise, NTB. Finally, $\oint$ indicates the operation of stacking the stages sequentially.

\begin{table}[]
  \centering
  \caption{Detailed configurations of Next-ViT variants.}
  \label{tab:configs}
  \resizebox{0.45\textwidth}{!}{
  \begin{tabular}{ccc|c|c|c}
  \toprule
  Stages                        & Output size                                         & Layers                                                          & Next-ViT-S                & Next-ViT-B            & Next-ViT-L        \\  \midrule
  \multirow{5}{*}{Stem}         & \multirow{5}{*}{$\displaystyle{\frac{H}{4}} \times \frac{W}{4} $}  & \multirow{5}{*}{\shortstack{Convolution \\ Layers}}             & \multicolumn{3}{c}{$\text{Conv}\ 3\times3, C=64, S=2$}                       \\  \cmidrule(l){4-6}
                                &                                                     &                                                                 & \multicolumn{3}{c}{$\text{Conv}\ 3\times3, C=32, S=1$}                       \\  \cmidrule(l){4-6}
                                &                                                     &                                                                 & \multicolumn{3}{c}{$\text{Conv}\ 3\times3, C=64, S=1$}                       \\  \cmidrule(l){4-6} 
                                &                                                     &                                                                 & \multicolumn{3}{c}{$\text{Conv}\ 3\times3, C=64, S=2$}                       \\  \midrule
  \multirow{4}{*}{Stage 1}      & \multirow{4}{*}{$\displaystyle{\frac{H}{4}} \times \frac{W}{4} $}  & \multirow{1}{*}{Patch Embedding}                                & \multicolumn{3}{c}{$\text{Conv}\ 1\times1, C=96$}                            \\  \cmidrule(l){3-6}
                                &                                                     & \multirow{2}{*}{\shortstack{Next-ViT \\ Block}}            & \multicolumn{3}{c}{\multirow{2}{*}{$\begin{bmatrix} \text{NCB} \times 3, 96 \end{bmatrix} \times 1$}}                    \\ 
                                &                                                     &                                                                 & \multicolumn{3}{c}{}                                                                                          \\  \midrule
  \multirow{6}{*}{Stage 2}      & \multirow{6}{*}{$\displaystyle{\frac{H}{8}} \times \frac{W}{8} $}  & \multirow{2.5}{*}{Patch Embedding}                              & \multicolumn{3}{c}{$\text{Avg\_pool}, S=2$}                                                                          \\  \cmidrule(l){4-6}
                                &                                                     &                                                                 & \multicolumn{3}{c}{$\text{Conv}\ 1\times1, C=192$}                                                                   \\  \cmidrule(l){3-6}
                                &                                                     & \multirow{3}{*}{\shortstack{Next-ViT \\ Block}}            & \multicolumn{3}{c}{\multirow{3}{*}{$\begin{bmatrix} \text{NCB} \times 3, 192 \\ \text{NTB} \times 1, 256 \end{bmatrix} \times 1$}}    \\
                                &                                                     &                                                                 & \multicolumn{3}{c}{}                                                                                          \\ 
                                &                                                     &                                                                 & \multicolumn{3}{c}{}                                                                                          \\  \midrule
  \multirow{6}{*}{Stage 3}      & \multirow{6}{*}{$\displaystyle{\frac{H}{16}} \times \frac{W}{16}$} & \multirow{2.5}{*}{Patch Embedding}                              & \multicolumn{3}{c}{$\text{Avg\_pool}, S=2$}                                                                          \\  \cmidrule(l){4-6}
                                &                                                     &                                                                 & \multicolumn{3}{c}{$\text{Conv}\ 1\times1, C=384$}                                                                   \\  \cmidrule(l){3-6}
                                &                                                     & \multirow{3}{*}{\shortstack{Next-ViT \\ Block}}            & \multirow{3}{*}{$\begin{bmatrix} \text{NCB} \times 4, 384 \\ \text{NTB} \times 1, 512 \end{bmatrix} \times 2$}        & \multirow{3}{*}{$\begin{bmatrix} \text{NCB} \times 4, 384 \\ \text{NTB} \times 1, 512 \end{bmatrix} \times 4$}        & \multirow{3}{*}{$\begin{bmatrix} \text{NCB} \times 4, 384 \\ \text{NTB} \times 1, 512 \end{bmatrix} \times 6$}       \\
                                &                                                     &                                                                 &                                                                                               &                                                                                               &                                                                                              \\ 
                                &                                                     &                                                                 &                                                                                               &                                                                                               &                                                                                              \\  \midrule    
  \multirow{6}{*}{Stage 4}      & \multirow{6}{*}{$\displaystyle{\frac{H}{32}} \times \frac{W}{32}$} & \multirow{2.5}{*}{Patch Embedding}                              & \multicolumn{3}{c}{$\text{Avg\_pool}, S=2$}                                                                          \\  \cmidrule(l){4-6}
                                &                                                     &                                                                 & \multicolumn{3}{c}{$\text{Conv}\ 1\times1, C=768$}                                                                   \\  \cmidrule(l){3-6}
                                &                                                     & \multirow{3}{*}{\shortstack{Next-ViT \\ Block}}            & \multicolumn{3}{c}{\multirow{3}{*}{$\begin{bmatrix} \text{NCB} \times 2,  \ 768 \\ \text{NTB} \times 1, 1024 \end{bmatrix} \times 1$}}    \\
                                &                                                     &                                                                 & \multicolumn{3}{c}{}                                                                                          \\ 
                                &                                                     &                                                                 & \multicolumn{3}{c}{}                                                                                          \\  \bottomrule 
  \end{tabular}
}
\vspace{-0.25cm}
\end{table}

\subsection{Next-ViT Architectures}
To provide a fair comparison with existing SOTA networks, we present three typical variants, namely, Next-ViT-S/B/L. The architecture specifications are listed in Table \ref{tab:configs}, in which $C$ represents output channel and $S$ denotes stride of each stage. Additionally, the channel shrink ratio $r$ in NTB is uniformly set as 0.75 and the spatial reduction ratio $s$ in E-MHSA is [8, 4, 2, 1] in different stages. The expansion ratios of MLP layer are set as 3 for NCB and 2 for NTB, respectively. The head dim in E-MHSA and MHCA is set as 32. For normalization layer and activation functions, both NCB and NTB use BatchNorm and ReLU.

% ImageNet-1K
\begin{table}[]
    \centering
    \caption{Comparison of different state-of-the-art methods on ImageNet-1K classification. HardSwish is not well supported by CoreML, $*$ denotes we replace it with GELU for fair comparison. $\dagger$ denotes we use large-scale dataset follow SSLD\cite{ssld}.}
    \label{tab:ImageNet_1K}
    \scalebox{0.65}{
    \begin{tabular}{l|c|cc|cc|c}
    \toprule
    \multirow{2}{*}{Method}         & Image    & Param     & FLOPs    & \multicolumn{2}{c|}{Latency(ms)}       & Top-1  \\ \cmidrule{5-6}
                                    & Size       & (M)       & (G)      & TensorRT     & CoreML       & (\%)       \\   \midrule
    ResNet101\cite{ResNet}       & 224    & 44.6      & 7.9       & 7.8      & 4.0   & 80.8          \\
    ResNeXt101-32x4d\cite{ResNeXt}       & 224    & 44.2  &8.0    & 8.0      & 4.0   & 78.8          \\
	RegNetY-8G\cite{RegNet}     & 224      & 44.2  &8.0      & 11.4      & 4.1    & 81.7          \\ 
	ResNeSt50\cite{zhang2022resnest}       & 224    & 27.5      & 5.4      & 102.7      & 36.6   & 81.1          \\
	EfficientNet-B3\cite{EfficientNet}       & 300    & 12.0      & 1.8      & 12.5       & 5.8   & 81.5          \\
    MobileViTv2-1.0\cite{mobilevit_v2}   &256   & 4.9      & 4.9       & -     & 2.9       & 78.1        \\
    MobileViTv2-2.0\cite{mobilevit_v2}   &256   & 18.5      & 7.5       & -     & 6.7          & 81.2        \\
	ConvNeXt-T\cite{ConvNext}      &224   & 29.0      & 4.5       & 19.0      &83.8    & 82.1          \\ 
    DeiT-T\cite{Deit}          &224    & 5.9      & 1.2       & 6.7   & 4.5  & 72.2        \\
    DeiT-S\cite{Deit}          &224    & 22.0      & 4.6       & 11.4   & 9.0    & 79.8        \\
    Swin-T\cite{Swin}          &224    & 29.0      & 4.5       & -   & -       & 81.3        \\
    PVTv2-B2\cite{PVT_v2}          &224    & 25.4    &4.0  &34.5    &  96.7     & 82.0        \\
    Twins-SVT-S\cite{Twins}     &224    & 24.0      & 2.9       & 17.3    & -   & 81.7        \\
    PoolFormer-S24\cite{metaformer}   &224   & 21.1      & 3.4       & 14.4     & 6.2      & 80.3        \\
    PoolFormer-S36\cite{metaformer}   &224   & 31.2      & 5.0       & 21.8     & 6.7     & 81.4        \\
    CMT-$\text{T}^*$\cite{Cmt}  &160 & 9.5      &0.6        & 11.8     & 4.6          & 79.1        \\ 
    CMT-$\text{XS}^*$\cite{Cmt} &192  & 15.2      & 1.5       & 21.9     & 8.3          & 81.8        \\
    CoaT Tiny\cite{CoaT}  &224 & 5.5      & 4.4       & 52.9     & 55.4          & 78.3        \\
    CvT-13\cite{CvT}    &224     & 20.1      & 4.5       & 18.1     & 62.6          & 81.6        \\
    % TRT-ViT-B \cite{xia2022trt}               & 43.1      & 4.0       & 5.6        & 4.0        & 82.1        \\
    \textbf{Next-ViT-S} &\textbf{224}    &\textbf{31.7}      &\textbf{5.8}       &\textbf{7.7}   &  \textbf{3.5}        & \textbf{82.5}        \\
    Next-ViT-S &384  & 31.7      & 17.3       & 21.6   &   8.9        & 83.6        \\
    Next-ViT-$\text{S}^\dagger$ &224  & 31.7      & 5.8       & 7.7   &   3.5        & 84.8        \\
    Next-ViT-$\text{S}^\dagger$ &384  & 31.7      & 17.3       & 21.6   &   8.9        & 85.8        \\
    \midrule
    ResNet152\cite{ResNet}       &224          & 60.2      & 4.0      & 11.3      & 5.0   & 81.7          \\
    ResNeXt101-64x4d\cite{ResNeXt}   & 224    & 83.5 &15.6     & 13.6      & 6.8   & 79.6          \\
    ResNeSt101\cite{zhang2022resnest}       &224      & 48.0      & 10.2      & 149.8      & 45.4   & 83.0          \\
    ConvNeXt-S\cite{ConvNext}      &224    & 50.0      & 8.7       & 28.1      &159.5    & 83.1          \\
    
    Swin-S\cite{Swin}            &224  & 50.0      & 8.7       & -   & -      & 83.0        \\
    PVTv2-B3\cite{PVT_v2}          &224    &45.2   &6.9  & 55.8    &  107.7      & 83.2        \\
    Twins-SVT-B\cite{Twins}        &224    & 56.0      & 8.6       & 32.0    & -     & 83.2        \\
    PoolFormer-M36\cite{metaformer}     &224  & 56.1      & 8.8       &28.2          &  -     & 82.1        \\
    CSWin-T\cite{CSWin}            &224     & 23.0      & 4.3       & 29.5     & -     & 82.7        \\
    CoaT Mini\cite{CoaT}          &224  & 10.0      & 6.8       & 68.0     & 60.8          & 81.0        \\
    CvT-21\cite{CvT}       &224     & 32.0      & 7.1       & 28.0     & 91.4          & 82.1        \\
    UniFormer-S\cite{uniformer}       &224  & 22.1      & 3.6       & 14.4         & 4.6      & 82.9       \\
    TRT-ViT-C\cite{xia2022trt}         &224  & 67.3      & 5.9       & 9.2         & 5.0      & 82.7        \\
    \textbf{Next-ViT-B}     &\textbf{224}  & \textbf{44.8}  & \textbf{8.3} & \textbf{10.5}    & \textbf{4.5}        & \textbf{83.2}        \\  
    Next-ViT-B     &384   & 44.8      & 24.6       & 29.6   & 12.4          & 84.3        \\
    Next-ViT-$\text{B}^\dagger$     &224   & 44.8      & 8.3       & 10.5   & 4.5          & 85.1        \\
    Next-ViT-$\text{B}^\dagger$     &384   & 44.8      & 24.6       & 29.6   & 12.4          & 86.1        \\
    \midrule 
    RegNetY-16G \cite{RegNet}        &224   & 84.0      & 16.0      & 18.0      & 7.4   & 82.9          \\
    EfficientNet-B5\cite{EfficientNet}        &456           & 30.0      & 9.9      & 64.4      & 23.2   & 83.7          \\
    ConvNeXt-B \cite{ConvNext}             &224        & 88.0      & 15.4       & 37.3      & 247.6    & 83.9          \\
    DeiT-B \cite{Deit}                 &224             & 87.0      & 17.5      & 31.0    & 18.2       & 81.8        \\
    Swin-B \cite{Swin}                 &224             & 88.0      & 15.4      & -    & -      & 83.3        \\
    PVTv2-B4\cite{PVT_v2}          &224    &62.6        &10.1   &70.8     &  139.8      & 83.6        \\
    Twins-SVT-L \cite{Twins}            &224            & 99.2      & 15.1      & 44.1    & -     & 83.7        \\
    PoolFormer-M48\cite{metaformer}          &224         & 73.2      & 11.6       &38.2          &  -     & 82.5        \\
    CSWin-S\cite{CSWin}                &224           & 35.0      & 6.9       & 45.0     & -     & 83.6        \\
    CMT-$\text{S}^*$\cite{Cmt}    &224  & 25.1      & 4.0       & 52.0     & 14.6          & 83.5        \\
    CoaT Small\cite{CoaT}  &224          & 22.0      & 12.6       & 82.7     & 122.4          & 82.1        \\
    UniFormer-B\cite{uniformer}   &224                & 50.2      & 8.3       & 31.0        & 9.0      & 83.9        \\
    TRT-ViT-D\cite{xia2022trt}     &224              & 103.0     & 9.7       & 15.1           &8.3     & 83.4        \\
    EfficientFormer-L7\cite{li2022efficientformer} &224 & 82.0      & 7.9       & 17.4     & 6.9          & 83.3        \\
    \textbf{Next-ViT-L}    &\textbf{224}       & \textbf{57.8}      & \textbf{10.8}      & \textbf{13.0}    & \textbf{5.5}         & \textbf{83.6}        \\  
    Next-ViT-L       &384         & 57.8      & 32.0      & 36.0    &15.2         & 84.7        \\
    Next-ViT-$\text{L}^\dagger$       &224         & 57.8      & 10.8      & 13.0    &5.5         & 85.4        \\
    Next-ViT-$\text{L}^\dagger$       &384         & 57.8      & 32.0      & 36.0    &15.2         & 86.4        \\
    \bottomrule
    \end{tabular}
    }
\end{table}

% semantic segmentation ADE20K
\begin{table*}[]
    \caption{Comparison of different backbones on ADE20K semantic segmentation task. FLOPs are measured with the input size of 512$\times$2048. $\dagger$ denotes training Semantic FPN-80K for 80k iteration with a total batch size of 32, which is 2$\times$ training data iteration compared to regular setting. $\dagger$ indicates that the model is pre-trained on large-scale dataset.}
    \centering
    \resizebox{0.9\textwidth}{!}{
    \begin{tabular}{l|cc|ccc|ccc}
    \toprule
    \multirow{2.5}{*}{Backbone}    &\multicolumn{2}{c|}{Latency(ms)}    & \multicolumn{3}{c|}{Semantic FPN 80k}   & \multicolumn{3}{c}{UperNet 160k}          \\ \cmidrule(l){2-9}
                                 & TensorRT  & CoreML    & Param(M)   & FLOPs(G)   & mIoU(\%)      & Param(M)  & FLOPs(G)    & mIoU/MS mIoU(\%)  \\   \midrule
    ResNet101\cite{ResNet}                    &32.8  &13.2    & 48.0       & 260        & 38.8          & 96          & 1029          & -/44.9            \\
    ResNeXt101-32x4d\cite{ResNeXt}   &37.0  & 15.3   & 47.1 &- & 39.7       & -         & -          & -/-            \\
    ResNeSt50\cite{zhang2022resnest}      &522.3  &372.3    & 30.4       & 204        & 39.7          & 68.4          & 973          & 42.1/-            \\
    ConvNeXt-T\cite{ConvNext}                  &78.5  &349.4    & -       & -        & -          & 60.0          & 939          & -/46.7            \\
    Swin-T\cite{Swin}            & - & -   & 31.9       & 182        & 41.5          & 59.9       & 945        & 44.5/45.8      \\
    PVTv2-B2\cite{PVT_v2}        &167.6  & 101.0   &29.1  &-        & 45.2          & -       & -        & -/-      \\
    Twins-SVT-S\cite{Twins}      &127.2  &-    & 28.3       & 144        & 43.2          & 54.4       & 901        & 46.2/47.1      \\
    PoolFormer-S12\cite{metaformer}      &30.3    &18.2        & 15.7   & -     & 37.2          & -          & -          & -/-            \\
    PoolFormer-S24\cite{metaformer}      &59.1    &23.0        & 23.2   & -     & 40.3          & -          & -          & -/-            \\
    TRT-ViT-$\text{C}^*$\cite{xia2022trt} &40.6  &20.7    & 70.6       & 213        & 46.2          & 105.0          & 978          & 47.6/48.9        \\
    EfficientFormer-L3\cite{li2022efficientformer}      &35.9  & 10.6   & -       & -        & 43.5          & -          & -          & -/-            \\
    \textbf{Next-ViT-S}          &\textbf{38.2}  &\textbf{18.1}      &\textbf{36.3}       & \textbf{208}        & \textbf{46.5}          & \textbf{66.3}      & \textbf{968}       & \textbf{48.1}/\textbf{49.0}      \\ \midrule
    ResNeXt101-64x4d\cite{ResNeXt}   &65.7  & 25.6   & 86.4 &- &40.2       & -          & -          & -/-            \\
    ResNeSt101\cite{zhang2022resnest}    &798.9  &443.6    & 51.2       & 305        & 42.4          & 89.2          & 1074          & 44.2/-            \\
    ConvNeXt-S\cite{ConvNext}                  &131.5  &658.0    & -       & -        & -          & 82.0          & 1027          & -/49.6            \\
    Swin-S\cite{Swin}            &-  &-    & 53.2       & 274        & 45.2          & 81.3       & 1038       & 47.6/49.5      \\
    PVTv2-B3\cite{PVT_v2}        &230.9  &  114.5  &49.0 &- &47.3          & -       & -        & -/-      \\
    Twins-SVT-B\cite{Twins}      &231.8  &-    & 60.4       & 261        & 45.3          & 88.5       & 1020       & 47.7/48.9      \\
    PoolFormer-S36\cite{metaformer}      &87.7    &27.7        & 34.6   & -     & 42.0          & -          & -          & -/-            \\
    PoolFormer-M36\cite{metaformer}      & 127.8   & 32.0       & 59.8  & -      & 42.4          & -          & -          & -/-            \\
    CSWin-T\cite{CSWin}          &182.3  & -   & 26.1       & 202        & 48.2          & 59.9       & 959        & 49.3/50.4      \\
    UniFormer-$\text{S}^*$\cite{uniformer} &90.7  &33.5    & 25.0       & 247        & 46.6          & 52.0     & 1088          & 47.6/48.5            \\
    TRT-ViT-$\text{D}^*$\cite{xia2022trt} &58.1  &29.4    & 105.9       & 296        & 46.7          & 143.7          & 1065          & 48.8/49.8            \\
    EfficientFormer-L7\cite{li2022efficientformer}      &84.0  &23.0    & -       & -        & 45.1          & -          & -          & -/-            \\
    \textbf{Next-ViT-B}          &\textbf{51.6}  & \textbf{24.4}      & \textbf{49.3}       &\textbf{260}        & \textbf{48.6}          & \textbf{79.3}      & \textbf{1020}       & \textbf{50.4}/\textbf{51.1}      \\   \midrule
    ConvNeXt-B\cite{ConvNext}                  &181.3  &1074.4    & -       & -        & -          & 122.0          & 1170          & -/49.9            \\       
    Swin-B\cite{Swin}            &-  &-    & 91.2       & 422        & 46.0          & 121.0      & 1188       & 48.1/49.7      \\
    PVTv2-B4\cite{PVT_v2}        &326.4 &  147.6  &66.3 &- &47.9          & -       & -        & -/-      \\
    Twins-SVT-L\cite{Twins}      &409.7  &-    & 103.7      & 404        & 46.7          & 133.0      & 1164       & 48.8/49.7      \\
    PoolFormer-M48\cite{metaformer}      &168.7    & 39.5       & 77.1  & -      & 42.7          & -          & -          & -/-            \\
    CSWin-S\cite{CSWin}          &298.2  & -    & 38.5       & 271        & 49.2          & 64.4       & 1027       & 50.4/51.5      \\
    UniFormer-$\text{B}^*$\cite{uniformer} &195.1  &70.6    & 54.0       & 471        & 48.0          & 80.0     & 1227          & 50.0/50.8            \\
    
    \textbf{Next-ViT-L}          &\textbf{65.3}  &\textbf{30.1}     &\textbf{62.4}       & \textbf{331}       & \textbf{49.1}          & \textbf{92.4}      & \textbf{1072}       & \textbf{50.1}/\textbf{50.8} \\
    \midrule
    Swin-$\text{B}^\dagger$\cite{Swin}            &-  &-    & -       & -        & -          & 121.0      & 1841       & 50.0/51.7      \\
    CSWin-$\text{B}^\dagger$\cite{Swin}            &-  &-    & -       & -        & -          & 109.2      & 1941       & 51.8/52.6      \\
    Next-ViT-$\text{S}^\dagger$          &38.2  &18.1      &36.3       & 208        & 48.8         & 66.3      & 968       & 49.8/50.8      \\
    Next-ViT-$\text{B}^\dagger$          &51.6  & 24.4      & 49.3       &260        & 50.2          & 79.3      & 1020       & 51.8/52.8 \\
    Next-ViT-$\text{L}^\dagger$          &65.3  &30.1     &62.4       & 331       & 50.5          & 92.4      & 1072      & 51.5/52.0 \\
    \bottomrule
    \end{tabular}}
    \label{tab:ADE20K}
\end{table*}
\section{Experimental Results}
\label{Experimental_section}

\subsection{ImageNet-1K Classification}
\subsubsection{Implementation} 
We carry out the image classification experiment on the ImageNet-1K \cite{ImageNet-1K}, which contains about 1.28M training images and 50K validation images from 1K categories.
For a fair comparison, we follow the training settings of the recent vision Transformer \cite{PVT_v1,Twins, sepvit, Scalablevit} with minor changes.
Concretely, all of the Next-ViT variants are trained for 300 epochs on 8 V100 GPUs with a total batch size of 2048. The resolution of the input image is resized to 224 $\times$ 224. 
We adopt the AdamW \cite{AdamW} as the optimizer with weight decay 0.1.
The learning rate is gradually decayed based on the cosine strategy with the initialization of 2e-3 and the use of a linear warm-up strategy with 20 epochs for all Next-ViT variants.
Besides, we have also employed the increasing stochastic depth augmentation \cite{Stochasticdepth} with the maximum drop-path rate of 0.1, 0.2, 0.2 for Next-ViT-S/B/L. Models with $\dagger$ are trained on large-scale dataset follow SSLD\cite{ssld}. For 384 $\times$ 384 input size, we fine-tune the models for 30 epochs with the weight decay of 1e-8, learning rate of 1e-5, batch size of 1024.
With the input size corresponding to the respective method, latency in Table \ref{tab:ImageNet_1K} is uniformly measured based on the TensorRT-8.0.3 framework with a T4 GPU (batch size=8) and CoreML framework on an iPhone12 Pro Max with iOS 16.0 (batch size=1). Note that both the iPhone 12 and iPhone 12 Pro Max are equipped with the same A14 processor.

% Mask R-CNN
\begin{table*}[]
    \caption{Comparison of different backbones on Mask R-CNN-based object detection and instance segmentation tasks. FLOPs are measured with the inpus size of $800 \times 1280$. The superscript $b$ and $m$ denote the box detection and mask instance segmentation.}
    \centering
    \resizebox{0.99\textwidth}{!}{
    \begin{tabular}{c|cc|cc|cccccc|cccccc}
    \toprule
    \multirow{2}{*}{Backbone}    &Param   & FLOPs     & \multicolumn{2}{c|}{Latency(ms)}     & \multicolumn{6}{c|}{Mask R-CNN 1$\times$}        & \multicolumn{6}{c}{Mask R-CNN 3$\times$ + MS}       \\ \cline{4-17} 
                                 &(M)   &(G)   &TensorRT     &CoreML        & AP$^b$        & AP$_{50}^b$   & AP$_{75}^b$   & AP$^m$        & AP$_{50}^m$   & AP$_{75}^m$   & AP$^b$        & AP$_{50}^b$   & AP$_{75}^b$   & AP$^m$        & AP$_{50}^m$   & AP$_{75}^m$   \\ \midrule
    ResNet101\cite{ResNet}      & 63.2      & 336   &32.8   &13.2      & 40.4          & 61.1          & 44.2          & 36.4          & 57.7          & 38.8          & 42.8          & 63.2          & 47.1          & 38.5          & 60.1          & 41.3          \\
    ResNeXt101-32x4d\cite{ResNeXt}      & 62.8      & -    &37.0   &  15.3    & 41.9 &62.5 &45.9 &37.5 &59.4 &40.2          & -          & -          & -          & -          & -          & -          \\
    ResNeSt50\cite{zhang2022resnest}      & 47.4      & 400   &522.3   &372.3      & 42.6          & -          & -          & 38.1          & -          & -          & -          & -          & -          & -          & -          & -          \\
    
    % RegNetX-3.2GF\cite{RegNet}      & 34.5      & 236.7   &22.7   & 6.54      & 40.3          & 61.6          & 44.1          & 36.6          & 58.5          & 39.4          & 43.1          & 63.8          & 47.0          & 38.7          & 60.5          & 41.8          \\
    % RegNetX-4.0GF\cite{RegNet}      & 41.2      & 253.0   &28.5   &7.38      & 41.5          & 62.7          & 45.7          & 37.4          & 59.4          & 40.3          & 43.4          & 64.0          & 47.8          & 39.2          & 61.3          & 42.1          \\
    % RegNetX-8.0GF\cite{RegNet}      & 58.2      & 335.7   &36.0   &12.3      & 41.7          & 62.7          & 46.1          & 37.5          & 59.6          & 39.9          & -          & -          & -          & -          & -          & -          \\
    ConvNext-T\cite{ConvNext}            & -      & 262    & 78.5  & 349.4     & -  & - & - & - & - & - &46.2 & 67.9& 50.8 &41.7 & 65.0 & 44.9          \\
    Swin-T\cite{Swin}            & 47.8      & 264    & -  & -     & 42.2          & 64.4          & 46.2          & 39.1          & 64.6          & 42.0          & 46.0          & 68.2          & 50.2          & 41.6          & 65.1          & 44.8          \\
    PVTv2-B2\cite{PVT_v2}            & 45.0      & -    & 167.6  & 101.0    & 45.3 &67.1 &49.6 &41.2 &64.2 &44.4          & -          & -          & -          & -          & -          & -          \\
    Twins-SVT-S\cite{Twins}      & 44.0      & 228   & 127.2  & -     & 43.4          & 66.0          & 47.3          & 40.3          & 63.2          & 43.4          & 46.8          & 69.2          & 51.2          & 42.6          & 66.3          & 45.8          \\
    PoolFormer-S12\cite{metaformer}       & 31.6    & -    & 30.3  & 18.2  &37.3   &59.0   &40.1   &34.6   &55.8   &36.9   &-  & -   &-    & -     &-     & -     \\
    PoolFormer-S24\cite{metaformer}       & 41.0    & -    &59.1   &22.9   &40.1   &62.2   &43.4   &37.0   &59.1   &39.6   &-  & -   &-    & -     &-     & -     \\
    TRT-ViT-C\cite{xia2022trt}       & 86.3      & 294    &40.6   &20.8     &44.7         & 66.9          &48.8         & 40.8         &63.9         & 44.0          &47.3         &68.8.   &51.9 &42.7           &65.9       &46.0        \\
    EfficientFormer-L3\cite{li2022efficientformer}     & -      & -   &35.9   &  10.6   & 41.4          & 63.9          & 44.7          & 38.1          & 61.0          & 40.4          & -          & -          & -          & -         & -          & -          \\
    \textbf{Next-ViT-S}         & \textbf{51.8}      & \textbf{290}   &\textbf{38.2}   &\textbf{18.1}      &\textbf{45.9}          &\textbf{68.3}          & \textbf{50.7}         & \textbf{41.8}          & \textbf{65.1}          & \textbf{45.1}          & \textbf{48.0}         & \textbf{69.7}          &\textbf{52.8}          &\textbf{43.2}          &\textbf{66.8}         & \textbf{46.7}          \\
     
    \midrule
    ResNeXt101-64x4d\cite{ResNeXt}      & 101.9      & -    &65.7   & 25.6     & 42.8 & 63.8 &47.3 &38.4 &60.6 &41.3          & -          & -          & -          & -          & -          & -          \\
    ResNeSt101\cite{zhang2022resnest}      & 68.1      & 499   &798.9   &443.6      & 45.2          & -          & -          & 40.2          & -          & -          & -          & -          & -          & -          & -          & -          \\
    Swin-S\cite{Swin}         & 69.1      & 354    & -   & -     & 44.8          & 66.6          & 48.9          & 40.9          & 63.4          & 44.2          & 48.5          & 70.2          & 53.5          & 43.3          & 67.3          & 46.6          \\
    PVTv2-B3\cite{PVT_v2}            & 64.9      & -    &230.9   &   114.5   & 47.0 &68.1 &51.7 &42.5 &65.7 &45.7          & -          & -          & -          & -          & -          & -          \\
    Twins-SVT-B\cite{Twins}    & 76.3      & 340   & 231.8  &   -   & 45.2          & 67.6          & 49.3          & 41.5          & 64.5          & 44.8          & 48.0          & 69.5          & 52.7          & 43.0          & 66.8          & 46.6          \\
    CSWin-T\cite{CSWin}          & 42.0      & 279   & 182.3  &  -    & 46.7          & 68.6          & 51.3          & 42.2          & 65.6          & 45.4          & 49.0          & 70.7          & 53.7          & 43.6          & 67.9          & 46.6          \\
    PoolFormer-S36\cite{metaformer}       & 50.5   & -    &87.7   & 27.7  &41.0   &63.1   &44.8   &37.7   &60.1   &40.0   &-  & -   &-    & -     &-     & -     \\
    
    CMT-S\cite{CSWin}       & 30.2      & -    &200.5   &73.4     & 44.6          & 66.8          & 48.9          & 40.7          & 63.9         &43.4         & -          & -          & -          & -       & -          & -           \\
    CoaT Mini\cite{CoaT}       & 41.6      & -    &509.6   &476.9     &45.1          &-         & -          & 40.6          &-         & -          & 46.5         &-         & -           & 41.8         &-       & -        \\
    UniFormer-S$_{h14}$\cite{uniformer}       & 41      & 269    &164.0   & -   &45.6   &68.1   &49.7   &41.6   &64.8   &45.0   &48.2   &70.4   &52.5   &43.4   &67.1   &47.0       \\
    TRT-ViT-D\cite{xia2022trt}       & 121.5      & 375    &58.1   & 29.5    &45.3   &67.9    &49.6  &41.6   &64.7   &44.8   &48.1   &69.3   &52.7   &43.4   &66.7   &46.8        \\
    EfficientFormer-L7\cite{li2022efficientformer}       & -      & -    & 84.0  & 23.0  & 42.6         & 65.1          & 46.1          & 39.0          & 62.2          & 41.7         & -          & - & -          & -       & -          & -           \\
    \textbf{Next-ViT-B}      &\textbf{64.9}   &\textbf{340}  &\textbf{51.6}   &\textbf{24.4}   &\textbf{47.2}   &\textbf{69.6}   &\textbf{51.6}   &\textbf{42.8}   &\textbf{66.5}   &\textbf{45.9}   &\textbf{49.5}   &\textbf{71.1}   &\textbf{54.2}   &\textbf{44.4}   &\textbf{68.3}  &\textbf{48.0}          \\

     \midrule
    Swin-B\cite{Swin}         & 107      & 496    & -  & -     & 46.9 & - & - & 42.3 & - & - & 48.5 & 69.8 & 53.2 & 43.4 & 66.8 & 46.9          \\
    PVTv2-B4\cite{PVT_v2}            & 82.2      & -    &326.4   & 147.6   & 47.5 &68.7 &52.0 &42.7 &66.1 &46.1          & -          & -          & -          & -          & -          & -          \\
    Twins-SVT-L\cite{Twins}    & 111      & 474   & 409.7  &   -   & 45.7          & -          & -          & 41.6          & -          & -          & -          &           & -          & -          & -          & -          \\
    CSWin-S\cite{CSWin}       & 54.0      & 342    &298.2   &-    & 47.9          & 70.1          & 52.6          & 43.2          & 67.1          & 46.2          & 50.0          & 71.3          & 54.7          & 44.5          & 68.4          & 47.7          \\
    CoaT Small\cite{CoaT}      & 54.0      & -    &601.7   &  612.9   & 46.5         &-         & -           & 41.8         &-         & -         & 49.0          &-         & -           &43.7         &-  & -        \\
    UniFormer-B$_{h14}$\cite{uniformer}       & 69      & 399    & 367.7  & -  &47.4   &69.7   &52.1   &43.1   &66.0   &46.5   &50.3   &72.7   &55.3   &44.8   &69.0   &48.3        \\
    
    \textbf{Next-ViT-L}                   &\textbf{77.9}   &\textbf{391}  &\textbf{65.3}   &\textbf{30.1}   &\textbf{48.0}   &\textbf{69.8}   &\textbf{52.6}   &\textbf{43.2}   &\textbf{67.0}   &\textbf{46.8}   &\textbf{50.2}   &\textbf{71.6}   &\textbf{54.9}   &\textbf{44.8}   &\textbf{68.7}   &\textbf{48.2}         \\

    \bottomrule
    \end{tabular}
    }

    \label{tab:MaskRCNN_COCO}
  \end{table*}

\subsubsection{Comparison with State-of-the-art Models} 
 As shown in Table \ref{tab:ImageNet_1K}, compared to the latest state-of-the-art methods (e.g. CNNs, ViTs and hybrid networks), we achieve the best trade-off between accuracy and latency.
%  Specifically, Next-ViT-S achieves 82.4\% top-1 accuracy, 1.6\% higher than ResNet101 \cite{ResNet}  with similar latency on TensorRT and 0.5ms faster on CoreML. Next-ViT-B achieves similar accuracy compared with recent SOTA models ConvNeXt-S \cite{ConvNext} while 1.7$\times$ faster on TensorRT. Meanwhile, base variant also achieves a SOTA inference speed on CoreML with 4.5ms.
%  Furthermore, Next-ViT-L costs about 60\% fewer inference time on TensorRT than CSWin-S \cite{CSWin} with same accuracy of 83.6.
%  When it comes to CoreML, compared to the recent models EfficientFormer-L7 \cite{li2022efficientformer}, Next-ViT-L also achieves better results while the inference speed is accelerated by 25\%. 
%  These results demonstrate that the proposed TensorRT-oriented Vision  Transformer design is an effective and promising paradigm.
Specifically, compared with the famous CNNs such as ResNet101 \cite{ResNet}, Next-ViT-S improves the accuracy by 1.7\% with a similar latency on TensorRT and faster speed on CoreML(from 4.0ms to 3.5ms). Meanwhile, Next-ViT-L achieves the similar accuracy as EfficientNet-B5 \cite{EfficientNet} and ConvNeXt-B while 4.0$\times$ and 
 1.4$\times$ faster on TensorRT, 3.2$\times$ and 44$\times$ faster on CoreML.
 In terms of the advanced ViTs, Next-ViT-S outperforms Twins-SVT-S \cite{Twins} by 0.8\% with 1.3$\times$ faster inference speed on TensorRT. Next-ViT-B surpasses CSwin-T\cite{CSWin} by 0.5\% while the inference latency is compressed by 64\% on TensorRT.
 Finally, compared with recent hybrid methods, Next-ViT-S beats CMT-XS by 0.7\% with 1.8$\times$ and 1.4$\times$ faster speed on TensorRT and CoreML. Compared to EfficientFormer-L7 \cite{li2022efficientformer}, Next-ViT-L predict with 20\% fewer runtime on CoreML and 25\% fewer runtime on TensorRT while the performance is improved from 83.3\% to 83.6\%. Next-ViT-L also obtains a 15\% inference latency gain and achieves a better performance than TRT-ViT-D.
 These results demonstrate that the proposed Next-ViT design is an effective and promising paradigm.

\subsection{ADE20K Semantic Segmentation}

\subsubsection{Implementation} 
To further verify the capacity of our Next-ViT, we conduct the semantic segmentation experiment on ADE20K \cite{ADE20K}, which contains about 20K training images and 2K validation images from 150 categories.
To make fair comparisons, we also follow the training conventions of the previous vision Transformers \cite{PVT_v1,Swin,Twins} on the Semantic FPN \cite{Semantic_FPN} and UperNet \cite{UperNet} frameworks. Most of models are pre-trained on the ImageNet-1k and models with $\dagger$ are pre-trained on large-scale dataset. All the models are pretrained with resolution 224$\times$224 and then trained on ADE20K with the input size of 512$\times$512. For the Semantic FPN framework, we adopt the AdamW optimizer with both the learning rate and weight decay being 0.0001. Then we train the whole network for 40K iterations with a total batch size of 32 based on the stochastic depth of 0.2 for Next-ViT-S/B/L.
For the training and testing on the UperNet framework, we also train the models for 160K iterations with the stochastic depth of 0.2. AdamW optimizer is used as well but with the learning rate $6\times10^{-5}$, total batch size 16, and weight decay 0.01. Then we test the mIoU based on both single-scale and multi-scale (MS) where the scale goes from 0.5 to 1.75 with an interval of 0.25. For detection and segmentation tasks, due to some modules in Mask R-CNN  and Upernet are not easy to be deployed on TensorRT and CoreML, we only measure the latency of the backbone for a fair comparison, with the same test environments as classification.  For simplicity, the input size of 512$\times$512 is uniformly  used to measure latency in Table \ref{tab:ADE20K} and Table \ref{tab:MaskRCNN_COCO}.

\subsubsection{Comparison with State-of-the-art Models} 
In Table \ref{tab:ADE20K}, we make a comparison with CNNs, ViTs, and recent hybrid methods as well. Next-ViT-S surpasses ResNet101~\cite{ResNet} and ResNeXt101-32x4d\cite{ResNeXt} by 7.7\% and 6.8\% mIoU, respectively. Next-ViT-B beats CSwin-T by 0.4\% mIoU and the inference speed is accelerated by 2.5$\times$ on TensorRT. Compared with the Uniformer-S/B~\cite{uniformer}, Next-ViT-B/L achieves 2.0\% and 1.1\% mIoU performance gain while 0.4$\times$/1.3$\times$ faster on CoreML and 0.8$\times$/1.6$\times$ faster on TensorRT. Next-ViT-B surpasses EfficientFormer-L7 \cite{li2022efficientformer} by 3.5\% mIoU with similar CoreML runtime and 38\% fewer latency on TensorRT.
In terms of the UperNet \cite{UperNet} framework, Next-ViT-S surpasses recent SOTA CNN model ConvNeXt\cite{ConvNext} 2.3\% MS mIoU while 1.0$\times$ and 18.0$\times$ faster on TensorRT and CoreML respectively. Compared to the CSWin-S\cite{CSWin}, Next-ViT-L achieves 3.6$\times$ faster speed on TensorRT with similar performance.
Extensive experiments reveal that our Next-ViT achieves excellent potential on segmentation tasks.

% Based on the Semantic FPN framework,  Next-ViT-S surpasses Twins-SVT-S \cite{Twins} by 3.8\% mIoU with about 83\% lower latency.
% Meanwhile, Next-ViT shows great advantage over CNNs (e.g., ResNet\cite{ResNet}, ResNeXt\cite{ResNeXt}). Specifically, Next-ViT-S achieves 44.2\% mIoU, which surpasses ResNet101 by 5.4\% but still faster(from 114ms to 82ms). 
% For the UperNet framework, Next-ViT-L achieves 2.2\% higher MS mIoU than recent SOTA CNN model ConvNeXt-T with 17\% fewer inference time. Next-ViT-L achieves similar accuracy with Twins-SVT-L while much faster (from 1632ms to 301ms). Extensive experiments reveal that our Next-ViT achieves excellent potential on segmentation tasks.

\subsection{Object Detection and Instance Segmentation}
\subsubsection{Implementation} 
Next, we evaluate Next-ViT on the objection detection and instance segmentation task \cite{COCO} based on the Mask R-CNN \cite{Mask_RCNN} frameworks with COCO2017 \cite{COCO}.
Specifically, all of our models are pre-trained on ImageNet-1K and then finetuned following the settings of the previous works \cite{PVT_v1,Swin,Twins}.
As for the 12 epochs (1$\times$) experiment, we use the AdamW optimizer with the weight decay of 0.05. There are 500 iterations for a warm-up during the training, and the learning rate will decline by 10$\times$ at epochs 8 and 11. Based on the 36 epochs (3$\times$) experiment with multi-scale (MS) training, models are trained with the resized images such that the shorter side ranges from 480 to 800 and the longer side is at most 1333. The learning rate will decline by 10$\times$ at epochs 27 and 33. The other settings are the same as 1$\times$.

\subsubsection{Comparison with State-of-the-art Models} 
Table \ref{tab:MaskRCNN_COCO} shows the evaluation results with the Mask R-CNN framework. Based on the 1$\times$ schedule, Next-ViT-S surpasses ResNet101\cite{ResNet} and ResNeSt50\cite{zhang2022resnest} by 5.5 AP$^b$ and 3.3 AP$^b$. Next-ViT-L beats PVTv2-B4\cite{PVT_v2} by 0.5 AP$^b$ and predict with 4.0$\times$ and 3.9$\times$ faster runtime on TensorRT and CoreML. Compared to the EfficientFormer-L7\cite{li2022efficientformer}, Next-ViT-B improves AP$^b$ from 42.6 to 47.2 with similar CoreML latency and 39\% fewer TensorRT runtime. Next-ViT-B outperforms TRT-ViT-D\cite{xia2022trt} by 1.9 AP$^b$  but is still faster on both TensorRT and CoreML.
Based on the 3$\times$ schedule, Next-ViT shows the same superiority as the 1$\times$. Specifically, Next-ViT-S surpasses ResNet101 by 5.2 AP$^b$ with similar latency. Compared to the Twins-SVT-S, Next-ViT-S achieves 1.2 AP$^b$ higher performance but with 3.2$\times$ faster speed on TensorRT. Next-ViT-B outperforms CSwin-T by 0.5 AP$^b$ but with 2.5$\times$ fewer prediction time.  For the Next-ViT-L, it exhibits a similar performance on object detection and instance segmentation as CSwin \cite{CSWin} but the inference speed is accelerated by 79\%.
% which also verify that Next-ViT achieves promising performance among these famous CNNs, ViTs, and hybrid methods.

\subsection{Ablation Study and Visualization}

\begin{figure}[t]
    \centering
    \includegraphics[width=0.47\textwidth]{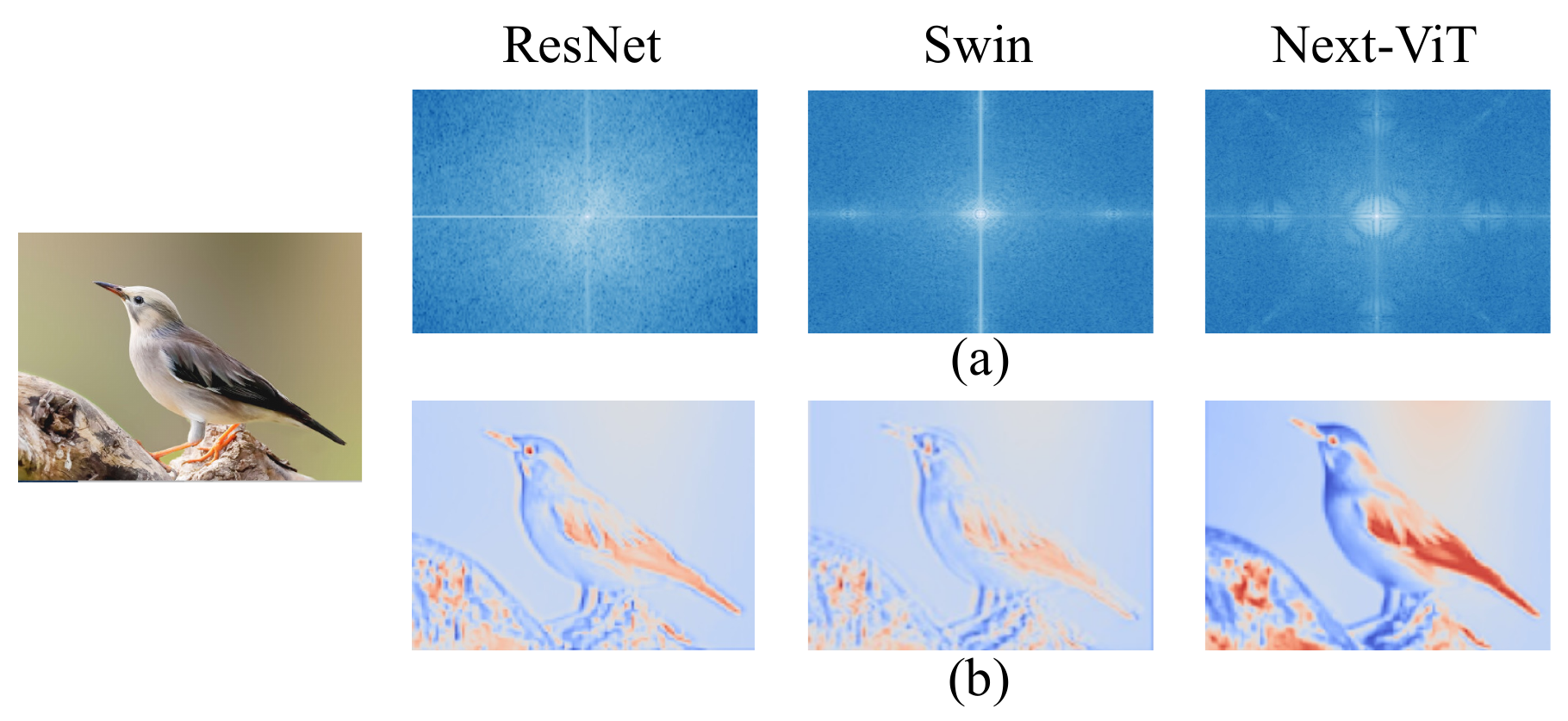}
    \caption{(a) Fourier spectrum of ResNet \cite{ResNet}, Swin \cite{Swin} and Next-ViT. (b) Heat maps of the output feature from ResNet \cite{ResNet}, Swin \cite{Swin} and Next-ViT.}
    \label{fig:visualization}
\end{figure}

	To understand our Next-ViT better, we ablate each critical design by evaluating its performance on ImageNet-1K classification and downstream tasks. We also visualize the Fourier spectrum and heat map of output features to show the intrinsic superiority of Next-ViT.
	
\subsubsection{Impact of Next Convolution Block}
 To verify the effectiveness of the proposed NCB, we replace NCB in Next-ViT with famous blocks, such as Bottleneck in ResNet \cite{ResNet}, ConvNeXt\cite{ConvNext} block, LSA block in Twins \cite{Twins}, and etc. For a fair comparison, we consistently use NTB and NHS to build different models under similar latency on TensorRT.
 
 As shown in Table~\ref{tab:block_comparsion}, NCB achieves the best latency/accuracy trade-off on all of the three tasks, which verifies the advantage of the proposed NCB. For example, NCB outperforms the recent ConvNeXt block\cite{ResNet} by 2.9\% in classification, 4.5 AP$^b$ in detection and 2.8\% mIoU in segmentation.

\subsubsection{Impact of Different Shrink Ratios in NTB}
Furthermore, we explore the effect of shrink ratio $r$ of Next Transformer Block on the overall performance of Next-ViT.
As stated in Table~\ref{tab:ratio_comparsion}, decreasing the shrinking ratio $r$, i.e. the number of channels in E-MHSA module, will reduce the model latency. Furthermore, model with $r=0.75$ and $r=0.5$ achieve better performance over model with pure Transformer ($r=1$). This denotes that fusing multi-frequency signals in a proper manner will enhance the model ability of representation learning.

Specially, model with $r=0.75$ achieve the best latency/accuracy trade-off. It outperforms baseline model $(r=1.0)$ with 0.4\%, 0.5 AP$^b$ and 1.0\% mIoU on classification, detection and segmentation while is more lightweight.
The above results indicate the effectiveness of the proposed NTB block.

\subsubsection{Impact of Normalization and Activation}
 We further study the impact of different normalization layers and activation functions in Next-ViT. As shown in Table~\ref{tab:norm_comparsion}, both the LN and GELU bring negligible performance improvement but with significantly higher inference latency on TensorRT. On the other hand, BN and ReLU achieve the best latency/accuracy trade-off on overall tasks. 
 Therefore, we uniformly use BN and ReLU in Next-ViT for efficient deployment in realistic industrial scenarios.
 
%  Specifically, Next-ViT-S with BN and ReLU achieves a performance of 82.4\% (-0.3\%) accuracy, 45.8 (-0.2) AP$^b$ and 47.0\% (-0.2\%) mIoU while accelerating the inference latency by 17\%, which is commendable in realistic industrial scenarios.

\begin{table}[]
    \centering\caption{Comparison of different convolution blocks.}
    \resizebox{0.45\textwidth}{!}{
        \begin{tabular}{c|c|c|c|c}
            \toprule
            \multirow{2.5}{*}{Block type}      & Latency(ms)       & Cls   & Det      & Seg      \\ \cmidrule(l){2-5}
                                                    & TensorRT              & Acc(\%)       & AP$^b$                        & mIoU(\%)    \\ \midrule
            BottleNeck Block\cite{ResNet}             & 7.8               & 81.9          & 44.5                          & 45.5          \\ 
            ConvNeXt Block\cite{ConvNext}       & 7.8               & 79.6          & 41.4                          & 43.7          \\
            LSA Block\cite{Twins}               & 8.4               & 78.2          & 38.7                          & 40.1          \\
            PoolFormer Block\cite{metaformer}    & 7.6               & 80.9          & 42.3                          & 44.0          \\
            Local MHRA Block\cite{uniformer}                & 7.8               & 80.5          & 42.4                          & 43.4          \\
            \textbf{NCB (ours)}                     & \textbf{7.7}        & \textbf{82.5}     & \textbf{45.9}       & \textbf{46.5}          \\ 

            \bottomrule                          
    \end{tabular}}
    \label{tab:block_comparsion}
\end{table}

\begin{table}
    \centering\caption{Comparison of results of different ratios.}
    \resizebox{0.35\textwidth}{!}{
        \begin{tabular}{c|c|c|c|c}
            \toprule
            \multirow{2.5}{*}{$r$}      & Latency(ms)      & Cls           & Det           & Seg      \\ \cmidrule(l){2-5}
                                          & TensorRT       & Acc(\%)       & AP$^b$         & mIoU(\%)    \\ \midrule
            0.0                         & 6.6               & 80.9          & 43.1                          & 42.4         \\
            0.25                        & 6.9               & 81.3          & 45.0                          & 45.2          \\
            0.50                        & 7.3               & 82.2          &45.4                          & 46.0          \\ 
            \textbf{0.75}             & \textbf{7.7}               & \textbf{82.5}     & \textbf{45.9}        &\textbf{46.5}         \\ 
            1.0                        & 8.2               & 82.1          & 45.4                         & 45.5          \\ 
            \bottomrule                          
    \end{tabular}}
    \label{tab:ratio_comparsion}
\end{table}

\begin{table}[]
    \centering\caption{Different normalizations and activations comparison.}
    \resizebox{0.43\textwidth}{!}{
        \begin{tabular}{c|c|c|c|c|c}
            \toprule
            \multirow{2.5}{*}{Norm}   & \multirow{2.5}{*}{Activation}    & Latency(ms)       & Cls   & Det      & Seg      \\ \cmidrule(l){3-6}
                                      &              & TensorRT              & Acc(\%)       & AP$^b$                        & mIoU(\%)    \\ \midrule
            LN              & GELU          & 9.3               & 82.7          & 46.0                          & 46.7          \\
            LN              & ReLU          & 9.1               & 82.7          & 46.0                          & 46.6          \\ 
            BN              & GELU          & 8.0               & 82.5          & 45.9                          & 46.6          \\
             \textbf{BN}     &  \textbf{ReLU}     & \textbf{7.7}    &  \textbf{82.5}      & \textbf{45.9}     & \textbf{46.5}          \\ 

            \bottomrule                          
    \end{tabular}}
    \label{tab:norm_comparsion}
\end{table}

\subsubsection{Visualization}
 To verify the superiority of our Next-ViT, we visualize the Fourier spectrum and heat maps of the output features from ResNet, Swin Transformer and Next-ViT in Figure \ref{fig:visualization} (a). 
 The spectrum distribution of ResNet denotes that convolution blocks tend to capture high-frequency signals while difficult to focus on low-frequency information. On the other hand, ViT experts in capturing low-frequency signals but ignore high-frequency signals. Finally, Next-ViT is capable of simultaneously capturing high-quality and multi-frequency signals, which shows the effectiveness of NTB.
 
 Furthermore, As shown in Figure \ref{fig:visualization} (b), we can see that Next-ViT can capture richer texture information and more accurate global information (e.g. edge shape) compared with ResNet and Swin, which shows the stronger modeling capability of Next-ViT.
 
%  further compared NCB with other convolution blocks, such as Bottleneck in ResNet \cite{ResNet}, LSA block in Twins \cite{Twins} etc. For a fair comparison,  we employ different types of blocks to stack the network under a similar latency on TensorRT.
%  As shown in Table \ref{tab:block_comparsion}, we can see that NCB achieves the best performance on all of the three tasks, which verifies the advantage of the proposed NCB. For example, NCB outperforms the famous bottleneck block\cite{ResNet} by 0.5\% in classification, 1.3 AP$^b$ in detection and 0.9\% mIoU in segmentation.

\section{Conclusion}
	\label{section:conclusion}
	In this paper, we present a family of Next-ViT that stacks efficient Next Convolution Block and Next Transformer Block in a novel strategy to build powerful CNN-Transformer hybrid architecture for efficient deployment on both mobile device and server GPU. 
	Experimental results demonstrate that Next-ViT achieves a state-of-the-art latency/accuracy trade-off across diverse visual tasks, such as image classification, object detection and semantic segmentation. We believe that our work builds a stable bridge between academic research and industrial deployment in terms of visual neural network design.
    We hope that our work will provide new insights and promote more research in neural network architecture design for realistic industrial deployment.

% 	directly treat the TensorRT and CoreML latency as a computational efficiency metric to guide the model design, which provides a viable solution to alleviate the gap between research and realistic industrial deployment.
% 	We conduct extensive experiments, propose three important components for next generation Vision Transformer network design, and propose . 

%%%%%%%%% REFERENCES
{\small
\bibliographystyle{ieee_fullname}
\bibliography{egbib}

\begin{thebibliography}{10}\itemsep=-1pt

\bibitem{bullier2001integrated}
Jean Bullier.
\newblock Integrated model of visual processing.
\newblock {\em Brain research reviews}, 36(2-3):96--107, 2001.

\bibitem{Mobile-Former}
Yinpeng Chen, Xiyang Dai, Dongdong Chen, Mengchen Liu, Xiaoyi Dong, Lu Yuan,
  and Zicheng Liu.
\newblock Mobile-former: Bridging mobilenet and transformer.
\newblock {\em arXiv preprint arXiv:2108.05895}, 2021.

\bibitem{Twins}
Xiangxiang Chu, Zhi Tian, Yuqing Wang, Bo Zhang, Haibing Ren, Xiaolin Wei,
  Huaxia Xia, and Chunhua Shen.
\newblock Twins: Revisiting the design of spatial attention in vision
  transformers.
\newblock {\em arXiv preprint arXiv:2104.13840}, 2021.

\bibitem{ssld}
Cheng Cui, Ruoyu Guo, Yuning Du, Dongliang He, Fu Li, Zewu Wu, Qiwen Liu,
  Shilei Wen, Jizhou Huang, Xiaoguang Hu, et~al.
\newblock Beyond self-supervision: A simple yet effective network distillation
  alternative to improve backbones.
\newblock {\em arXiv preprint arXiv:2103.05959}, 2021.

\bibitem{CoAtNet}
Zihang Dai, Hanxiao Liu, Quoc~V Le, and Mingxing Tan.
\newblock Coatnet: Marrying convolution and attention for all data sizes.
\newblock {\em Advances in Neural Information Processing Systems},
  34:3965--3977, 2021.

\bibitem{CSWin}
Xiaoyi Dong, Jianmin Bao, Dongdong Chen, Weiming Zhang, Nenghai Yu, Lu Yuan,
  Dong Chen, and Baining Guo.
\newblock Cswin transformer: A general vision transformer backbone with
  cross-shaped windows.
\newblock {\em arXiv preprint arXiv:2107.00652}, 2021.

\bibitem{ViT}
Alexey Dosovitskiy, Lucas Beyer, Alexander Kolesnikov, Dirk Weissenborn,
  Xiaohua Zhai, Thomas Unterthiner, Mostafa Dehghani, Matthias Minderer, Georg
  Heigold, Sylvain Gelly, et~al.
\newblock An image is worth 16x16 words: Transformers for image recognition at
  scale.
\newblock {\em arXiv preprint arXiv:2010.11929}, 2020.

\bibitem{Cmt}
Jianyuan Guo, Kai Han, Han Wu, Chang Xu, Yehui Tang, Chunjing Xu, and Yunhe
  Wang.
\newblock Cmt: Convolutional neural networks meet vision transformers.
\newblock {\em arXiv preprint arXiv:2107.06263}, 2021.

\bibitem{Mask_RCNN}
Kaiming He, Georgia Gkioxari, Piotr Doll{\'a}r, and Ross Girshick.
\newblock Mask r-cnn.
\newblock In {\em Proceedings of the IEEE International Conference on Computer
  Vision}, pages 2961--2969, 2017.

\bibitem{ResNet}
Kaiming He, Xiangyu Zhang, Shaoqing Ren, and Jian Sun.
\newblock Deep residual learning for image recognition.
\newblock In {\em Proceedings of the IEEE Conference on Computer Vision and
  Pattern Recognition}, pages 770--778, 2016.

\bibitem{PiT}
Byeongho Heo, Sangdoo Yun, Dongyoon Han, Sanghyuk Chun, Junsuk Choe, and
  Seong~Joon Oh.
\newblock Rethinking spatial dimensions of vision transformers.
\newblock In {\em Proceedings of the IEEE/CVF International Conference on
  Computer Vision}, pages 11936--11945, 2021.

\bibitem{Mobilenets}
Andrew~G Howard, Menglong Zhu, Bo Chen, Dmitry Kalenichenko, Weijun Wang,
  Tobias Weyand, Marco Andreetto, and Hartwig Adam.
\newblock Mobilenets: Efficient convolutional neural networks for mobile vision
  applications.
\newblock {\em arXiv preprint arXiv:1704.04861}, 2017.

\bibitem{DenseNet}
Gao Huang, Zhuang Liu, Laurens Van Der~Maaten, and Kilian~Q Weinberger.
\newblock Densely connected convolutional networks.
\newblock In {\em Proceedings of the IEEE conference on computer vision and
  pattern recognition}, pages 4700--4708, 2017.

\bibitem{Stochasticdepth}
Gao Huang, Yu Sun, Zhuang Liu, Daniel Sedra, and Kilian~Q Weinberger.
\newblock Deep networks with stochastic depth.
\newblock In {\em European Conference on Computer Vision}, pages 646--661,
  2016.

\bibitem{kauffmann2014neural}
Louise Kauffmann, Stephen Ramano{\"e}l, and Carole Peyrin.
\newblock The neural bases of spatial frequency processing during scene
  perception.
\newblock {\em Frontiers in integrative neuroscience}, 8:37, 2014.

\bibitem{Semantic_FPN}
Alexander Kirillov, Ross Girshick, Kaiming He, and Piotr Doll{\'a}r.
\newblock Panoptic feature pyramid networks.
\newblock In {\em Proceedings of the IEEE/CVF Conference on Computer Vision and
  Pattern Recognition}, pages 6399--6408, 2019.

\bibitem{uniformer}
Kunchang Li, Yali Wang, Junhao Zhang, Peng Gao, Guanglu Song, Yu Liu, Hongsheng
  Li, and Yu Qiao.
\newblock Uniformer: Unifying convolution and self-attention for visual
  recognition.
\newblock {\em arXiv preprint arXiv:2201.09450}, 2022.

\bibitem{sepvit}
Wei Li, Xing Wang, Xin Xia, Jie Wu, Xuefeng Xiao, Min Zheng, and Shiping Wen.
\newblock Sepvit: Separable vision transformer.
\newblock {\em arXiv preprint arXiv:2203.15380}, 2022.

\bibitem{li2022efficientformer}
Yanyu Li, Geng Yuan, Yang Wen, Eric Hu, Georgios Evangelidis, Sergey Tulyakov,
  Yanzhi Wang, and Jian Ren.
\newblock Efficientformer: Vision transformers at mobilenet speed.
\newblock {\em arXiv preprint arXiv:2206.01191}, 2022.

\bibitem{COCO}
Tsung-Yi Lin, Michael Maire, Serge Belongie, James Hays, Pietro Perona, Deva
  Ramanan, Piotr Doll{\'a}r, and C~Lawrence Zitnick.
\newblock Microsoft coco: Common objects in context.
\newblock In {\em European Conference on Computer Vision}, pages 740--755,
  2014.

\bibitem{Swin}
Ze Liu, Yutong Lin, Yue Cao, Han Hu, Yixuan Wei, Zheng Zhang, Stephen Lin, and
  Baining Guo.
\newblock Swin transformer: Hierarchical vision transformer using shifted
  windows.
\newblock {\em arXiv preprint arXiv:2103.14030}, 2021.

\bibitem{ConvNext}
Zhuang Liu, Hanzi Mao, Chao-Yuan Wu, Christoph Feichtenhofer, Trevor Darrell,
  and Saining Xie.
\newblock A convnet for the 2020s.
\newblock {\em arXiv preprint arXiv:2201.03545}, 2022.

\bibitem{AdamW}
Ilya Loshchilov and Frank Hutter.
\newblock Decoupled weight decay regularization.
\newblock {\em arXiv preprint arXiv:1711.05101}, 2017.

\bibitem{ShuffleNet_v2}
Ningning Ma, Xiangyu Zhang, Hai-Tao Zheng, and Jian Sun.
\newblock Shufflenet v2: Practical guidelines for efficient cnn architecture
  design.
\newblock In {\em Proceedings of the European Conference on Computer Vision},
  pages 116--131, 2018.

\bibitem{MobileViT}
Sachin Mehta and Mohammad Rastegari.
\newblock Mobilevit: light-weight, general-purpose, and mobile-friendly vision
  transformer.
\newblock {\em arXiv preprint arXiv:2110.02178}, 2021.

\bibitem{mobilevit_v2}
Sachin Mehta and Mohammad Rastegari.
\newblock Separable self-attention for mobile vision transformers.
\newblock {\em arXiv preprint arXiv:2206.02680}, 2022.

\bibitem{park2022vision}
Namuk Park and Songkuk Kim.
\newblock How do vision transformers work?
\newblock {\em arXiv preprint arXiv:2202.06709}, 2022.

\bibitem{RegNet}
Ilija Radosavovic, Raj~Prateek Kosaraju, Ross Girshick, Kaiming He, and Piotr
  Doll{\'a}r.
\newblock Designing network design spaces.
\newblock In {\em Proceedings of the IEEE/CVF Conference on Computer Vision and
  Pattern Recognition}, pages 10428--10436, 2020.

\bibitem{ImageNet-1K}
Olga Russakovsky, Jia Deng, Hao Su, Jonathan Krause, Sanjeev Satheesh, Sean Ma,
  Zhiheng Huang, Andrej Karpathy, Aditya Khosla, Michael Bernstein, et~al.
\newblock Imagenet large scale visual recognition challenge.
\newblock {\em International Journal of Computer Vision}, 115(3):211--252,
  2015.

\bibitem{MobileNet_v2}
Mark Sandler, Andrew Howard, Menglong Zhu, Andrey Zhmoginov, and Liang-Chieh
  Chen.
\newblock Mobilenetv2: Inverted residuals and linear bottlenecks.
\newblock In {\em Proceedings of the IEEE Conference on Computer Vision and
  Pattern Recognition}, pages 4510--4520, 2018.

\bibitem{BoTNet}
Aravind Srinivas, Tsung-Yi Lin, Niki Parmar, Jonathon Shlens, Pieter Abbeel,
  and Ashish Vaswani.
\newblock Bottleneck transformers for visual recognition.
\newblock In {\em Proceedings of the IEEE/CVF conference on computer vision and
  pattern recognition}, pages 16519--16529, 2021.

\bibitem{EfficientNet}
Mingxing Tan and Quoc Le.
\newblock Efficientnet: Rethinking model scaling for convolutional neural
  networks.
\newblock In {\em International Conference on Machine Learning}, pages
  6105--6114, 2019.

\bibitem{Deit}
Hugo Touvron, Matthieu Cord, Matthijs Douze, Francisco Massa, Alexandre
  Sablayrolles, and Herv{\'e} J{\'e}gou.
\newblock Training data-efficient image transformers \& distillation through
  attention.
\newblock In {\em International Conference on Machine Learning}, pages
  10347--10357, 2021.

\bibitem{vaswani2017attention}
Ashish Vaswani, Noam Shazeer, Niki Parmar, Jakob Uszkoreit, Llion Jones,
  Aidan~N Gomez, {\L}ukasz Kaiser, and Illia Polosukhin.
\newblock Attention is all you need.
\newblock {\em Advances in neural information processing systems}, 30, 2017.

\bibitem{PVT_v2}
Wenhai Wang, Enze Xie, Xiang Li, Deng-Ping Fan, Kaitao Song, Ding Liang, Tong
  Lu, Ping Luo, and Ling Shao.
\newblock Pvtv2: Improved baselines with pyramid vision transformer.
\newblock {\em arXiv preprint arXiv:2106.13797}, 2021.

\bibitem{PVT_v1}
Wenhai Wang, Enze Xie, Xiang Li, Deng-Ping Fan, Kaitao Song, Ding Liang, Tong
  Lu, Ping Luo, and Ling Shao.
\newblock Pyramid vision transformer: A versatile backbone for dense prediction
  without convolutions.
\newblock {\em arXiv preprint arXiv:2102.12122}, 2021.

\bibitem{CvT}
Haiping Wu, Bin Xiao, Noel Codella, Mengchen Liu, Xiyang Dai, Lu Yuan, and Lei
  Zhang.
\newblock Cvt: Introducing convolutions to vision transformers.
\newblock In {\em Proceedings of the IEEE/CVF International Conference on
  Computer Vision}, pages 22--31, 2021.

\bibitem{xia2022trt}
Xin Xia, Jiashi Li, Jie Wu, Xing Wang, Mingkai Wang, Xuefeng Xiao, Min Zheng,
  and Rui Wang.
\newblock Trt-vit: Tensorrt-oriented vision transformer.
\newblock {\em arXiv preprint arXiv:2205.09579}, 2022.

\bibitem{UperNet}
Tete Xiao, Yingcheng Liu, Bolei Zhou, Yuning Jiang, and Jian Sun.
\newblock Unified perceptual parsing for scene understanding.
\newblock In {\em European Conference on Computer Vision}, pages 418--434,
  2018.

\bibitem{ResNeXt}
Saining Xie, Ross Girshick, Piotr Doll{\'a}r, Zhuowen Tu, and Kaiming He.
\newblock Aggregated residual transformations for deep neural networks.
\newblock In {\em Proceedings of the IEEE Conference on Computer Vision and
  Pattern Recognition}, pages 1492--1500, 2017.

\bibitem{CoaT}
Weijian Xu, Yifan Xu, Tyler Chang, and Zhuowen Tu.
\newblock Co-scale conv-attentional image transformers.
\newblock In {\em Proceedings of the IEEE/CVF International Conference on
  Computer Vision}, pages 9981--9990, 2021.

\bibitem{Scalablevit}
Rui Yang, Hailong Ma, Jie Wu, Yansong Tang, Xuefeng Xiao, Min Zheng, and Xiu
  Li.
\newblock Scalablevit: Rethinking the context-oriented generalization of vision
  transformer.
\newblock {\em arXiv preprint arXiv:2203.10790}, 2022.

\bibitem{metaformer}
Weihao Yu, Mi Luo, Pan Zhou, Chenyang Si, Yichen Zhou, Xinchao Wang, Jiashi
  Feng, and Shuicheng Yan.
\newblock Metaformer is actually what you need for vision.
\newblock In {\em Proceedings of the IEEE/CVF Conference on Computer Vision and
  Pattern Recognition}, pages 10819--10829, 2022.

\bibitem{T2T}
Li Yuan, Yunpeng Chen, Tao Wang, Weihao Yu, Yujun Shi, Zi-Hang Jiang,
  Francis~EH Tay, Jiashi Feng, and Shuicheng Yan.
\newblock Tokens-to-token vit: Training vision transformers from scratch on
  imagenet.
\newblock In {\em Proceedings of the IEEE/CVF International Conference on
  Computer Vision}, pages 558--567, 2021.

\bibitem{zhang2022resnest}
Hang Zhang, Chongruo Wu, Zhongyue Zhang, Yi Zhu, Haibin Lin, Zhi Zhang, Yue
  Sun, Tong He, Jonas Mueller, R Manmatha, et~al.
\newblock Resnest: Split-attention networks.
\newblock In {\em Proceedings of the IEEE/CVF Conference on Computer Vision and
  Pattern Recognition}, pages 2736--2746, 2022.

\bibitem{Shufflenet}
Xiangyu Zhang, Xinyu Zhou, Mengxiao Lin, and Jian Sun.
\newblock Shufflenet: An extremely efficient convolutional neural network for
  mobile devices.
\newblock In {\em Proceedings of the IEEE Conference on Computer Vision and
  Pattern Recognition}, pages 6848--6856, 2018.

\bibitem{ADE20K}
Bolei Zhou, Hang Zhao, Xavier Puig, Sanja Fidler, Adela Barriuso, and Antonio
  Torralba.
\newblock Scene parsing through ade20k dataset.
\newblock In {\em Proceedings of the IEEE Conference on Computer Vision and
  Pattern Recognition}, pages 633--641, 2017.

\bibitem{FAN}
Daquan Zhou, Zhiding Yu, Enze Xie, Chaowei Xiao, Animashree Anandkumar, Jiashi
  Feng, and Jose~M Alvarez.
\newblock Understanding the robustness in vision transformers.
\newblock In {\em International Conference on Machine Learning}, pages
  27378--27394. PMLR, 2022.

\end{thebibliography}
}

\end{document}